\documentclass[sigconf,screen]{acmart}
\usepackage[english]{babel}
\usepackage{blindtext}
\usepackage[subtle, title=normal]{savetrees}
\usepackage{multirow}
\usepackage{fvextra}
\usepackage{subcaption}
\usepackage{enumitem}
\usepackage{xcolor}
\usepackage{listingsutf8}   
\usepackage{upquote}        
\usepackage[most]{tcolorbox}
\usepackage{tikz}
\usetikzlibrary{positioning,calc}
\usepackage{algpseudocode}
\usepackage{cleveref}

\makeatletter
\renewcommand{\@date}{} 
\makeatother
\makeatletter
\renewcommand{\sectionautorefname}{\S\@gobble}
\makeatother 
\makeatletter
\renewcommand{\subsectionautorefname}{\S\@gobble}
\makeatother 
\makeatletter
\renewcommand{\subsubsectionautorefname}{\S\@gobble} 
\makeatother

\AtBeginDocument{%
  \providecommand\BibTeX{{%
    \normalfont B\kern-0.5em{\scshape i\kern-0.25em b}\kern-0.8em\TeX}}}

\makeatletter
\renewcommand{\sectionautorefname}{\S\@gobble}
\makeatother 
\makeatletter
\renewcommand{\subsectionautorefname}{\S\@gobble}
\makeatother 
\makeatletter
\renewcommand{\subsubsectionautorefname}{\S\@gobble} 
\makeatother

\lstdefinestyle{pythonstyle}{
  language=Python,
  basicstyle=\ttfamily\scriptsize,
  keywordstyle=\color{blue},
  commentstyle=\color{teal},
  stringstyle=\color{brown},
  showstringspaces=false,
  breaklines=true,
  frame=single,
  numbers=none,
  tabsize=2,
  columns=fullflexible
}
\DeclareCaptionType{listing}[List of Listings][Listing] 
\usepackage{cuted}

\newcommand{\TheSystem}{Glia\xspace} 
\newcommand{\name}{\TheSystem} 

\newcommand{\eg}{{\em e.g., }}

\newcommand{\Fig}[1]{Fig.~\ref{fig:#1}}
\newcommand{\NewPara}[1]{\noindent{\bf #1}}

\newcommand{\eat}[1]{}
\newcommand{\Tab}[1]{Tab.~\ref{tab:#1}\xspace}

\setlist[itemize]{nosep,left=1pt}
\setlist[enumerate]{noitemsep, topsep=2pt, parsep=0pt, partopsep=0pt, leftmargin=*}

\lstdefinestyle{commandstyle}{
    language=Python,
    backgroundcolor=\color{gray!10},
    keywordstyle=\color{blue},
    stringstyle=\color{purple},
    commentstyle=\color{green!50!black},
    basicstyle=\ttfamily\scriptsize,
    breaklines=true,
    frame=single,
    framerule=0pt,
    framesep=5pt,
    rulecolor=\color{black},
    numbersep=5pt
}

\lstdefinestyle{bashstyle}{
    language=bash,
    backgroundcolor=\color{gray!10},
    keywordstyle=\color{blue},
    stringstyle=\color{purple},
    commentstyle=\color{green!50!black},
    basicstyle=\ttfamily\scriptsize,
    breaklines=true,
    frame=single,
    framerule=0pt,
    framesep=5pt,
    rulecolor=\color{black},
    numbersep=5pt
}

\lstdefinestyle{outputstyle}{
    backgroundcolor=\color{blue!5},
    basicstyle=\ttfamily\scriptsize,
    frame=tblr, 
    framesep=8pt,
    breaklines=true,
}

\lstdefinestyle{conversationstyle}{
    basicstyle=\sffamily\small,   
    backgroundcolor=\color{green!10},
    breaklines=true,
    frame=tblr,                   
    framerule=1pt,                
    rulecolor=\color{blue!60},    
    xleftmargin=3pt,             
    framexleftmargin=10pt,        
    breakindent=0pt,
}

\usepackage{pgfplotstable}
\usepackage{tikz}
\usetikzlibrary{fit,backgrounds,decorations.pathreplacing}
\usepgfplotslibrary{groupplots,statistics,fillbetween}
\pgfplotsset{compat=1.18}

\usetikzlibrary{
  shapes.misc, shapes.geometric, shapes.multipart, shapes.arrows,
  positioning, quotes, arrows.meta, decorations.pathmorphing, 
  matrix, graphs, fit, calc, automata, shadows
}

\DeclareGraphicsExtensions{.pdf,.png,.jpg}
\usepackage{epstopdf}
\usepackage{varwidth}

\usepackage{caption}
\AtBeginDocument{%
  \providecommand\BibTeX{{%
    Bib\TeX}}}

\acmSubmissionID{}

\setcopyright{none}
\renewcommand{\footnotetextcopyrightpermission}[1]{}

\begin{document}
\title{Glia: A Human-Inspired AI for Automated Systems Design and Optimization}


\author{
Pouya Hamadanian\textsuperscript{*$\ddagger$}\quad 
Pantea Karimi\textsuperscript{*$\ddagger$}\quad
Arash Nasr-Esfahany\textsuperscript{*$\ddagger$}\quad 
Kimia Noorbakhsh\textsuperscript{*$\ddagger$}\quad
Joseph Chandler\textsuperscript{$\ddagger$}\quad
Ali ParandehGheibi\textsuperscript{\S}\quad
Mohammad Alizadeh\textsuperscript{$\ddagger$}\quad
Hari Balakrishnan\textsuperscript{$\ddagger$}\\[12pt]
\textsuperscript{*}Equal contribution \quad
\textsuperscript{$\ddagger$}MIT CSAIL\quad
\textsuperscript{\S}Independent Researcher
}


\renewcommand{\shortauthors}{}

\begin{abstract}
    Can AI autonomously design mechanisms for computer systems on par with the creativity and reasoning of human experts? 
We present {\bf \name}, an AI architecture for networked systems design that uses large language models (LLMs) in a human-inspired multi-agent workflow. 
Each agent specializes in reasoning, experimentation, and analysis, collaborating through an evaluation framework that grounds abstract reasoning in empirical feedback.
Unlike prior ML-for-systems methods that optimize black-box policies, \name generates interpretable designs and exposes its reasoning. 
When applied to a distributed GPU cluster for LLM inference, it produces new algorithms for request routing, scheduling, and auto-scaling that perform at human-expert levels in significantly less time, while yielding novel insights into workload behavior. 
Our results suggest that combining reasoning LLMs with structured experimentation, an AI can produce creative and understandable designs for complex systems problems.

\end{abstract}

\maketitle
\pagestyle{plain}











\section{Introduction}
\label{s:intro}

{\em Can we develop an AI to tackle the design and optimization of networked systems and produce solutions on par with, or even surpass, those of PhD-level system engineers?} 

This question motivates our work. It matters because:
\begin{enumerate}

\item {\bf AI may soon be essential to manage complexity.} Modern computer systems are extraordinarily complex, driven by massive scale, rapidly evolving hardware, dynamic workloads, stringent performance demands, and escalating costs. Many organizations struggle to keep pace, with engineers stretched thin to deliver improvements. On the research front, understanding intricate system interactions takes longer than ever, slowing innovation. We believe today’s systems have reached a level of sophistication that makes traditional, human-centric R\&D insufficient for sustained progress.

\item {\bf The need for design velocity in the age of AI.} A central systems challenge of our time is building infrastructure capable of efficiently delivering AI applications. These workloads consume enormous computing resources, process vast amounts of data, and often operate under strict latency  constraints. While AI models advance at a breathtaking pace, the systems that support them lag behind~\cite{vllm,sarathi,Li2024LLMServingSurvey}, creating a growing gap between what AI demands and what current infrastructure can supply. 

\item {\bf Intellectual curiosity and the nature of design itself.} Given the remarkable progress of AI in many domains~\cite{AlphaCode, mit2023chatgpt,brachman2025currentfutureuselarge,deepmind2024gemini}, can it also produce good system designs? Hallmarks of good design such as simplicity, clarity, and robustness~\cite{wheeler_psd,wikiquote_dijkstra,harvard_design_principles} are difficult to quantify, unlike objective performance metrics. Our goal is thus twofold: to test whether an AI can generate high-performing designs that optimize specific performance objectives, and to examine whether its designs are understandable and insightful in the way that human-engineered designs often are.

\end{enumerate}

For over a decade, researchers have explored the use of AI and machine learning (AI/ML) for systems~\cite{switchml,network_planning_RL,295531,10.1145/3387514.3405859,Mao2017Pensieve,Khani2020Adaptive,salman2018deepconf,Marcus2019Neo,Li2023Reparo,Marcus2021Bao}. 
Yet despite hundreds of papers, real-world deployments remain rare. As we discuss in \S\ref{sec:related}, these approaches have often been {\em incomprehensible}, e.g., relying on opaque neural policies often developed with reinforcement learning (RL)~\cite{Mao2017Pensieve,Mao2016Resource,Mao2019Scheduling,10.1145/3387514.3405859}, and {\em fragile}, failing outside their training regimes~\cite{246358,jay2019internetcongestioncontroldeep,pcc,mao2020realworldvideoadaptationreinforcement,daneshamooz2025addressingmldomainadaptation}. Researchers and practitioners alike have thus had little confidence that such systems will behave reliably in unforeseen situations. There is, therefore, good reason to be skeptical of the real-world impact of efforts in AI/ML for systems.

But this time may be different. Large Language Models (LLMs) now excel at code generation, dramatically accelerating software development. They can ingest and synthesize vast amounts of text, identifying what matters and what does not, a hallmark of ``understanding.'' They can also solve mathematical and analytical problems, including those requiring symbolic reasoning and logical synthesis.  

Learning from prior ML-for-systems experience, our goal for the proposed AI is {\em not} to merely produce the ``best possible'' design on a fixed set of benchmarks or traces. Instead, our aim is to {\em build an AI that generates system designs and insights that impress human experts}. Performance improvements are important and a welcome by-product. Good designs and optimizations are clear and explainable, can be stress-tested and analyzed, and adapt to new situations, reducing the risk of unpleasant surprises.

One solution might be to craft detailed prompts for existing LLMs to produce system designs. As we show in \S\ref{s:llmalone}, however, this approach does not work well. 
Fundamentally, systems research integrates four interdependent skills: 
(a) developing and reasoning about a {\em model} of the system and problem, 
(b) formulating {\em hypotheses} about bottlenecks and designing {\em experiments} to test them, 
(c) {\em instrumenting and analyzing} telemetry data that capture performance and diagnostic metrics, and 
(d) {\em synthesizing insights} from these analyses into improved designs. 
These skills operate in a feedback loop, where an engineer iterates until satisfied with the outcome (or gives up). 
Moreover, systems research is inherently {\em collaborative}: teams combine complementary skills, vet ideas, and refine them through critique and iteration.

We introduce \name, {\em an AI architecture inspired by this human process}. 
It comprises three components: 
(1) a {\em front-end} for humans to specify tasks and provide background; 
(2) a {\em multi-agent AI} composed of LLM-based agents with reasoning, summarization, and analytical capabilities, each specialized for particular tasks and able to explore ideas sequentially or concurrently; and 
(3) an {\em evaluation framework} that could be a simulator, emulator, or testbed for running experiments and generating data for the AI agents to reason about. 

We apply \name to a distributed GPU-cluster system serving LLM inference requests. 
It produces new insights in designing workload-specific adaptive algorithms for 
(a) {\em request routing} (deciding which GPU should serve a request), 
(b) {\em batch scheduling} (dispatching batches of requests to individual GPUs running inference tasks), and 
(c) {\em auto-scaling} (adjusting cluster size dynamically to meet latency goals while controlling compute costs).

The key contributions and findings of this paper are:

\begin{enumerate}
    
\item {\bf Human-inspired AI design:} 
\name employs an agentic workflow that mirrors how expert humans design systems, including conceptual understanding, hypothesis formation, experimental testing, ideation, and iterative refinement. We compare this approach with prior methods such as AlphaEvolve~\cite{AlphaEvolve}. For example, on a benchmark developing a request router for a distributed LLM inference system serving a challenging workload, \name produced a novel routing algorithm whose mean request completion time is 2.2$\times$ lower than the  standard least-loaded queue (LLQ) baseline and 1.6$\times$ lower than the solution produced by OpenEvolve (an open-source implementation of AlphaEvolve). \name took only two hours to match the performance of a human expert who took two weeks to achieve a similar result.

\item {\bf Demonstration of \name's creativity:} 
\name autonomously generated novel resource management algorithms that are simple and interpretable, revealing the reasoning that led to their creation. These outputs provided new insights even to experts. For instance, \name discovered within one hour that poor performance in a particular LLM workload stemmed not from load imbalance, as initially assumed, but from a memory management bottleneck---an insight that took a human expert several days to uncover independently.

\item {\bf Workload-specific adaptability:} 
\name operates continuously, adapting to changes in workload and environment. When the workload characteristics shift, previously discovered algorithms may degrade in performance. In such cases, \name is able to generate new, more effective methods tailored to the new conditions.

\item {\bf Benefits of parallel multi-context execution:} 
Running \name with multiple concurrent contexts yields higher-quality solutions than a single one-shot execution. Parallel exploration prevents the system from converging prematurely on suboptimal ideas and encourages diversity in solution strategies.

\end{enumerate}

In \S\ref{sec:related}, we survey prior work on AI/ML for systems and recent relevant advances in LLMs.
In \S\ref{s:llmservingcase}, we dive into a case study of LLM serving on a GPU cluster to analyze why black-box LLM prompting alone is insufficient to produce robust or interpretable system designs. This section also shows the results of running \name on the problem, highlighting the differences from prior approaches. We then present the design of \name in \S\ref{sec:design}, contrasting it with systems such as AlphaEvolve~\cite{AlphaEvolve} and FunSearch~\cite{FunSearch}. Our evaluation in \S\ref{sec:eval} demonstrates \name's effectiveness using several experiments on the same case study. Finally, we reflect on our findings and discuss open challenges for agentic AI in systems design in \S\ref{sec:conclusion}.
\section{Related Work}
\label{sec:related}

\paragraph{AI/ML for networked systems.} Remy uses an RL-inspired approach to synthesize congestion control algorithms offline~\cite{winstein2013remy}, while PCC Vivace~\cite{dong2018pcc} and Aurora~\cite{jay2019aurora} use online learning and deep RL. For video streaming, Pensieve learns ABR policies that surpass human heuristics~\cite{Mao2017Pensieve}. In datacenters, researchers have applied RL to traffic optimization and control (e.g., AuTO~\cite{chen2018auto}, Iroko~\cite{ruffy2018iroko}), topology/routing management (DeepConf~\cite{salman2018deepconf}), and packet classification (NeuroCuts~\cite{liang2019neurocuts}). Despite impressive benchmarks, these approaches rely on simulators that miss key real-world artifacts or narrow traces, yielding opaque or complex policies with fragile generalization outside the training regime~\cite{puffer}.

DeepRL has also been applied to cluster scheduling (DeepRM~\cite{mao2016deeprm}; Decima~\cite{mao2019decima}), cache replacement (LeCaR~\cite{vietri2018lecar}), GPU warp scheduling (RLWS~\cite{anantpur2017rlws}), databases (CDBTune~\cite{zhang2019cdbtune}, Bao~\cite{Marcus2021Bao}), and datacenter cooling~\cite{lazic2018dccooling}. These efforts show that machine learning can find high-performance policies, but they often produce black-box artifacts that are hard to analyze, verify, or adapt under workload shifts. 

In practice, RL agents try many poor algorithms before finding a good one, and if something changes about the system, such as the workload, objective, configuration, or hardware, they need to reoptimize all over again. By contrast, \name 
decomposes the task into modeling, hypothesis generation, experiment design, and telemetry analysis; iterates with an evaluation-in-the-loop; and produces interpretable, stress-testable designs rather than a trained policy.

\paragraph{LLM-based discovery.}
With the rise of LLMs, much research has focused on using them for algorithm discovery~\cite{liu2024llm4ad, xie2025far}. These methods go beyond using LLMs for simple code generation and instead propose approaches that combine reasoning and search. Gottweis et al.~\cite{gottweis2025towards} employ multiple agents that collaborate through several iterations to refine and improve solutions. Other works, such as Evolution of Heuristics~\cite{EoH}, ShinkaEvolve~\cite{lange2025shinkaevolve}, AlphaEvolve~\cite{Novikov2025AlphaEvolve}, MCTS~\cite{zheng2025monte}, LAS~\cite{liu2025fitness}, and X-evolve~\cite{zhai2025x}, develop evolutionary search approaches in which populations of programs are evolved through mutation, crossover, and selection based on a fitness score. Recently, Multi-Objective Evolution of Heuristics~\cite{yao2025multi} has gained attention. Some works, such as CALM~\cite{huang2025calm}, combine evolutionary search with fine-tuning or employ supervised fine-tuning on curated datasets to enhance reasoning for algorithm discovery~\cite{liu2025fine}. Recent work from Google~\cite{aygun2025ai} proposes an agentic, tree-search-based approach. DeepEvolve augments AlphaEvolve with deep web research to make sure the idea behind each code corresponds with latest ideas in literature~\cite{liu2025scientific}. SR-Scientist \cite{xia2025sr} introduces a long-horizon symbolic regression framework where an LLM iteratively refines equations using external tools for data analysis and evaluation. Unlike Glia, the LLM cannot perform these analyses autonomously and depends on pre-defined tools for these capabilities. Thus, many of these approaches rely on black-box exploration, as discussed in \S\ref{sec:design}.

\paragraph{LLM-based heuristic design.}
Researchers have been exploring LLM-based algorithms to improve the performance of various systems. AI-Driven Research for Systems (ADRS)~\cite{cheng2025barbariansgateaiupending} is a recent effort that studies a class of evolutionary, LLM-centered frameworks (e.g., OpenEvolve~\cite{openevolve}, GEPA~\cite{agrawal2025gepa}, ShinkaEvolve~\cite{lange2025shinkaevolve}) across performance problems in networking, databases, and distributed systems. 
Shypula et al.~\cite{shypula2025automated} explore the ability of LLMs to optimize C++ code. Wei et al.~\cite{wei2024improving} introduce an Agent-System Interface (ASI), composed of a concise Domain-Specific Language (DSL) and a feedback interpreter called AutoGuide, to optimize mapper code for parallel programs. Nagda et al.~\cite{nagda2025reinforced} use AlphaEvolve to discover new finite combinatorial constructions in complexity theory. Sun et al.~\cite{sun2025automatically} employ LLMs to automatically discover heuristics for SAT solvers. Press et al.~\cite{press2025algotune} introduce AlgoTune, an agentic framework for optimizing general-purpose numerical programs. Liu et al.~\cite{liu2025alphago} propose ASI-ARCH, a closed-loop, multi-agent evolutionary system for neural architecture search. Astra~\cite{wei2025astra} and GPU Kernel Scientist~\cite{andrews2025gpu} explore the optimization of CUDA GPU kernels. NADA~\cite{10.1145/3696348.3696868} investigates the use of LLMs for designing adaptive bitrate streaming (ABR) algorithms. Robusta~\cite{karimi2025robusta} combines combinatorial reasoning about heuristics from prior work~\cite{10.1145/3696348.3696884,metaease} with LLM-based reasoning in an evolutionary search to discover networking heuristics with improved worst-case guarantees. He et al.~\cite{he2025congestioncontroloptimizationlarge} explores the use of LLMs in congestion control. POLICYSMITH \cite{dwivedula2025man} also explores LLM heuristic search for web caching and congestion control. MetaMuse \cite{ma2025algorithm} guides LLMs to generate novel algorithms with structured creative ideation, exploring the solution space using external stimuli, waypoint reasoning, and feedback-based performance embeddings.

\name differs from prior work in several ways:
\begin{enumerate}
\item It focuses on generating hypotheses and ideas from performance analysis rather than code evolution, which helps establish causal relationships to observed outcomes.
\item It analyses experimental results and modifies code based on the results of this analysis.
\item It proposes and runs new experiments to understand deeper relationships between instrumented metrics (not just the objective) and the bottlenecks.
\end{enumerate}

\section{Case Study: LLM Serving in a GPU Cluster}
\label{s:llmservingcase}

In this section we apply \name to the problem of efficiently serving inference requests across a cluster of GPUs running large language models (LLMs), which is a fundamental challenge in distributed LLM serving systems~\cite{orca2022yu,vllm,sarathi,splitwise,sglang,dynamo}. 
When an inference request arrives, the system must process it promptly and in a cost-efficient manner. 
Because GPUs are expensive and often operate under diverse workloads, achieving high utilization without violating latency constraints is crucial. 
As model architectures, hardware generations, and application workloads evolve, maintaining and optimizing serving systems across these combinations has become increasingly difficult. 

Today's systems implement one or two generic mechanisms to route requests to GPUs (discussed below). 
They also employ standard batch schedulers to dispatch requests within a GPU and may use auto-scalers to adjust cluster capacity as load fluctuates over time. 
However, these components are often tuned conservatively or manually, leaving significant performance and cost improvements unrealized. 
Optimizing them requires expert knowledge of both workloads and system internals, requiring specialized engineering teams to design and tune these systems~\cite{orca2022yu,vllm,splitwise}.

We focus on a key opportunity for optimization: {\em dynamic, workload-adaptive routing of inference requests}. 
The goal is to tailor the routing to observed workload patterns to best satisfy specified service-level objectives (SLOs). 
Typical SLOs include the mean {time to first token} (TTFT), which captures latency; the mean {\em time per output token} (TPOT), which is a measure of throughput; and the mean {\em end-to-end request completion time}, which reflects overall responsiveness.

\begin{figure}
    \centering
    \includegraphics[width=\linewidth]{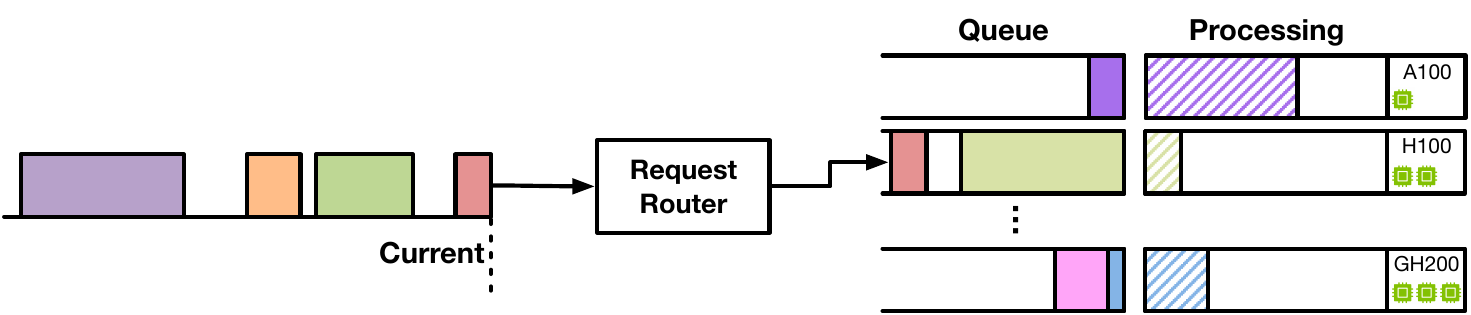}
    \caption{Pipeline of request routing for LLM inference.}
    \label{fig:routing}
\end{figure}

Request routing, illustrated in \Fig{routing}, is implemented in the {\em orchestration layer} of the inference stack. 
Modern LLM-serving systems such as NVIDIA Dynamo~\cite{dynamo}, Red Hat \texttt{llm-d}~\cite{llmd}, the vLLM production stack~\cite{production_stack}, and ByteDance AIBrix~\cite{aibrix} typically use simple, static policies such as:

\begin{itemize}
    \item \textbf{Round-robin (RR)}: routes each request to the next inference engine in sequence.
    \item \textbf{Least-loaded queue (LLQ)}: routes each request to the inference engine with the fewest queued requests.
    \item \textbf{Prefix-aware routing:} routes requests sharing a prefix to the same inference engine to enable reuse of cached computations; this technique is often combined with other criteria such as queue length or memory availability.\footnote{We disable prefix-aware routing in our experiments to focus on the core routing logic.}
\end{itemize}

While effective under steady conditions, these heuristics do not adapt to changing workloads or resource states. 
We therefore examine the potential for dynamic, workload-adaptive optimization of the request router using \name.

\subsection{Using LLMs As-is}
\label{s:llmalone}

\begin{figure}
    \centering
    \includegraphics[width=0.9\linewidth]{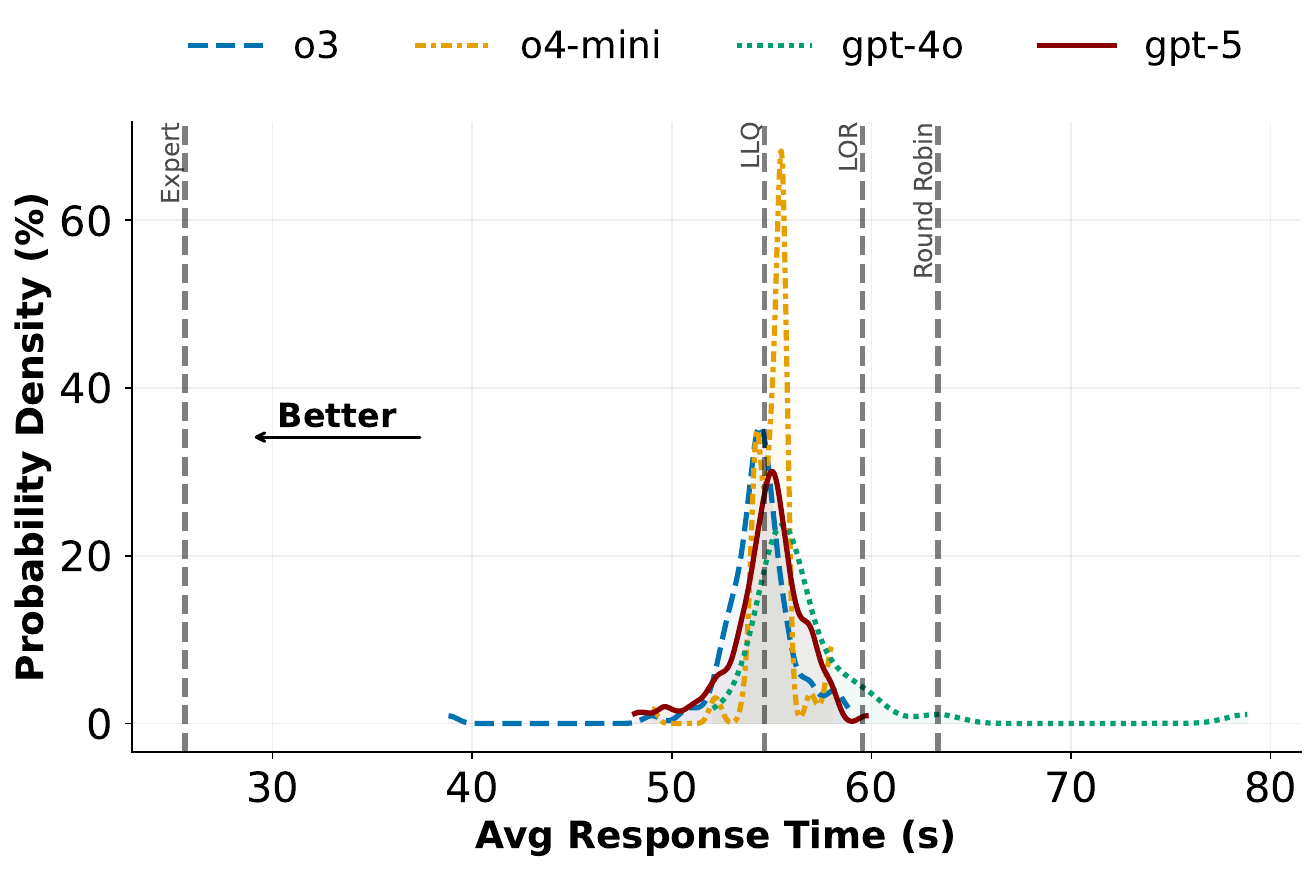}
    \vspace{-5 pt}
    \caption{Distribution of mean request completion times for 100 programs generated by directly prompting the LLM.}
    \label{fig:llmalonefigure}
    \vspace{-10 pt}
\end{figure}

A natural first step is to ask an LLM to {\em write the algorithm} given a detailed prompt describing the problem, environment, workload, and objectives. 
Even with carefully constructed prompts and state-of-the-art reasoning models, the resulting solutions are not competitive out of the box for this specialized task. 
\Fig{llm_as_is_prompt} shows an example of such a detailed prompt. 
\Fig{llmalonefigure} shows the distribution of mean request completion times for 100 generated programs sampled from the same prompt using \texttt{o3}, \texttt{o4-mini}, \texttt{gpt-4o}, and \texttt{gpt-5}. 
Performance varies widely across model outputs and is consistently worse than that of a human expert, indicating that direct prompting alone is insufficient for generating efficient request routing algorithms.

\subsection{Black-box LLM-in-the-loop Search}
\label{sec:motivation:monkey}

A more sophisticated approach places LLMs within a {\em black-box search loop}. 
In this setting, one or more LLMs generate or modify code candidates, an evaluator executes each on a benchmark and returns a performance score (\eg latency or throughput), and the LLM refines subsequent candidates based on that feedback~\cite{FunSearch,AlphaEvolve,EoH}. 
\Fig{funsearch_prompt} shows a typical prompt for use with systems such as FunSearch~\cite{FunSearch}.

This paradigm has some advantages: because proposals are in the form of code, thousands of variants can be generated and evaluated in parallel; i.e., the evolutionary loop can explore a large design space. 
However, these methods treat the problem largely as {\em code-level mutation and optimization}. 
Systems such as FunSearch~\cite{FunSearch} and AlphaEvolve~\cite{AlphaEvolve} operate directly on program text, mutating or recombining snippets without developing explicit hypotheses or structured reasoning traces. 
Consequently, ``idea evolution'' is driven by low-level code edits rewarded by a scalar objective, with little feedback or analysis about {\em why} a candidate works or fails. 
Few mechanisms exist to extract or leverage higher-level design insights—the kind that human experts rely on when developing new ideas. Many of the code candidates don't make logical sense, but yet the system experiments with them and obtains scores. This approach resembles a ``code monkey'' that tries out code variants, instead of reasoning about the ideas at a better level of abstraction~\cite{ehrlich2025codemonkeysscalingtesttimecompute}.

\Fig{funsearch-code} shows an example request-routing implementation generated by FunSearch. 
The resulting algorithm is a weighted composite cost function over a handful of input signals, a representation more typical of RL or ML-derived policies than of human-engineered designs grounded in analytical reasoning. 
Such solutions offer limited interpretability, depend on ad hoc weight calibration, and are often sensitive to small workload or configuration changes. 
As we show in \S\ref{sec:eval}, this black-box approach performs worse than the more explainable and adaptive designs produced by \name.

The core limitation is not the LLM’s reasoning capability or the evolutionary framework itself, but the {\em level of abstraction} at which the system operates: reasoning purely in code, with limited visibility into system behavior or design principles.

\subsection{The \name Approach}
\label{sec:motivation:sys}

Rather than using LLM as a black-box optimizer, \name uses it to elicit {\em systems reasoning}. We design an agentic LLM framework that mirrors how human engineers approach design problems—a workflow loop that forms hypotheses, conducts experiments, analyzes results, and refines ideas.
The system is named {\em \name}, inspired by glial cells in the brain. Just as glial cells support, maintain, and enhance the function of neurons, \name supports, augments, and accelerates the work of systems researchers and engineers.

This reasoning-centered methodology is {\em generalizable} across system domains and aims to be {\em robust}, avoiding the false paths of brute-force or code-mutation methods. By combining learned expertise with continuous experimentation, \name produces algorithms that are explainable, adaptable to changing workloads, and grounded in analytic reasoning rather than opaque optimization.

We illustrate Glia’s operation in the context of the request-routing problem. The user specifies a task prompt, an example of which is in \Fig{glia_task_prompt} to the \name front-end. Starting from a baseline such as LLQ, \name iterates through cycles of experimentation, analysis, idea generation, and code modification, progressively refining its design toward improved performance and clarity.

\paragraph{Experimentation.}  \name begins by testing a few simple request-routing strategies. 
It first evaluates the baseline performance of LLQ, then explores alternatives such as Least Outstanding Requests (LOR) and modified versions of LLQ where the load signal incorporates memory pressure. 
It executes each algorithm on the benchmark, recording detailed statistics and performance metrics for every run. 
Across these initial experiments, \name observes that none of these simple variants yield significant improvement over the baseline.

\paragraph{Analysis.} After about ten such experiments, \name determines that further blind exploration is unproductive:
\begin{tcolorbox}[
  colback=green!5!white,  
  colframe=green!75!black, 
  rounded corners,          
  before skip=6pt, after skip=6pt
]
\small
Rather than continuing to iterate blindly, I am going to step back, profile the baseline more rigorously and derive a quantitative model of where time is actually spent (queueing vs execution, pre-emptions, GPU-memory pressure, ..). Once that is done we can target the dominant contributor directly and should be able to reach the desired 30\% speed-up.
\end{tcolorbox}

\name proceeds to instrument the simulator and collect detailed internal metrics. From this analysis, it identifies {\em restarts}—instances where a request is evicted and restarted—as a potential source of inefficiency, an insight discovered autonomously without user prompting. 
\name prints the following summary:

\begin{lstlisting}[style=outputstyle]
mean e2e no restart 37.40
mean e2e restarted 48.12
fraction restarted 0.26
\end{lstlisting}

\name finds that roughtl;y 26\% of the requests in this workload experience restarts, with restarted requests averaging 48.1\,s end-to-end latency—about 11\,s slower than non-restarted ones. \name formulates the following hypothesis:
\begin{tcolorbox}[
  colback=green!5!white,  
  colframe=green!75!black, 
  rounded corners,          
]
\small
Hypothesis: memory imbalance $\rightarrow$ more restarts $\rightarrow$ wasted execution time.
\end{tcolorbox}

Modern GPUs are highly parallel devices designed to process many requests concurrently. To achieve efficiency, they batch requests together, making scheduling decisions critical to performance. LLM inference workloads are often bursty, with highly skewed and unpredictable compute demands. Request durations vary with prefill and decode token lengths, frequently following heavy-tailed distributions~\cite{yang2024queueingtheoreticperspectivelowlatency,tao2025promptawareschedulinglowlatencyllm}. 

A core scheduling challenge is that each request’s GPU memory requirement is proportional to its eventual output length, which is unknown at routing time. A router must either allocate the maximum possible size~\cite{orca2022yu} or allocate incrementally, assuming the batch fits in available GPU memory~\cite{vllm}. The latter, used in vLLM, avoids over-provisioning but risks out-of-memory conditions. When memory is exhausted during batch processing, the youngest request is evicted, freeing its memory but losing all progress. This eviction triggers a restart and wastes computation.

\noindent \paragraph{\name discovers a solution.} Having identified restarts as a major source of inefficiency, \name , \name experiments with several techniques from the literature to mitigate them. It reports that it has ``implemented several increasingly sophisticated global-scheduler prototypes: memory-head-room LB, prompt-token balanced LB, elephant-house heuristic, and size-affinity batching LB.'' 
Its evaluation of these approaches in the simulator finds that the best method so far, {\em least outstanding blocks}, improves average latency by only 3\% over the LLQ baseline. It says that it is ``continuing to iterate on schedulers''.

At this point, Glia's {\em Supervisor} agent intervenes:

\begin{tcolorbox}[
  colback=blue!5!white,    
  colframe=blue!75!black,  
  rounded corners,         
  before skip=6pt, after skip=6pt
]
\small
It seems that we have been unable to reduce the number of restarts. Why is that? Is there something structurally wrong about our approach that is making it difficult to limit the number of restarts?
\end{tcolorbox}

This step helps the Researcher agent achieve a breakthrough by triggering the following chain-of-thought reasoning:
\begin{tcolorbox}[
  colback=green!5!white,  
  colframe=green!75!black, 
  rounded corners,          
  before skip=6pt, after skip=6pt
]
\small
\textbf{Why restarts stubbornly persist}
\begin{enumerate}[leftmargin=0.5em, itemsep=0.1em]
    \item vLLM’s \textbf{pre-empt-on-growth} behaviour
    \begin{itemize}[leftmargin=0.5em, itemsep=0.1em]
        \item vLLM allocates just enough KV-cache blocks for the prompt.
        \item During decode it keeps adding blocks \textbf{1 token at a time} (1/16 of a block).
        \item As soon as a new block is needed and no free block is available, vLLM pre-empts the \textbf{youngest} request, frees its blocks and pushes it back to the pending queue.
        \item \textbf{\textit{This happens even if the replica was perfectly balanced at admission time.}}
    \end{itemize}
    \item The global scheduler sees \textbf{only prompt-length}, not future decode length.
    \begin{itemize}[leftmargin=0.5em, itemsep=0.1em]
        \item A 50-token prompt that later decodes 1500 tokens will eventually need [1500 / 16] = 94 extra blocks.
        \item The scheduler cannot foresee this growth, so an apparently ``safe'' admission can still trigger a pre-emption many seconds later.
    \end{itemize}
    \item Memory stays \textbf{near 100\% utilisation} most of the time
    \begin{itemize}[leftmargin=0.5em, itemsep=0.1em]
        \item LLQ (and our variants) keep GPUs busy to maximise throughput.
        \item With no head-room, the *first* request that needs an extra block causes a domino of restarts.
        \item Our heuristics moved load around, but average utilisation stayed > 90\%. That is simply too tight.
    \end{itemize}
\end{enumerate}
\end{tcolorbox}

Guided by this analysis, \name proposes a new routing strategy that explicitly forecasts memory usage (i.e., expected output tokens) and dispatches requests only when sufficient GPU headroom remains:

\begin{tcolorbox}[
  colback=green!5!white,  
  colframe=green!75!black, 
  rounded corners,          
]
\small
\textbf{To cut both queueing delay and pre-emptions we must}:
\begin{enumerate}[leftmargin=0.5em, itemsep=0.1em]
    \item Balance \textbf{memory usage / outstanding prompt tokens} across replicas, not just queue length.
    \item Admit new requests to replicas that still have KV-cache head-room so that pre-emptions become rare.
\end{enumerate}
\end{tcolorbox}

\name formalizes this idea by introducing two tunable parameters: the {\em decode-to-prefill ratio} $r$ and a {\em memory safety margin}, $m$. If the dynamically estimated $r$ for the workload exceeds a threshold and available memory falls below $(1-m)$ of total capacity, the router holds the request in a global queue rather than dispatching it immediately to a GPU. Intuitively, when $r$ is large, the eventual number of output tokens may exhaust GPU memory, and deferring admission mitigates this risk when the amount of available memory is small.

\name implements this strategy (\Fig{single-context-glia}), calling it the {\em Head-Room Allocator} (HRA), and evaluates it. The initial implementation performs worse than baseline LLQ (mean latency $>50$\,s vs.\ 40\,s) but eliminates most restarts, reducing them from 26\% to under 0.001\%. Recognizing the opportunity for parameter tuning, \name performs a rapid search over the $(r, m)$ parameter space, soon identifying a configuration that breaks the 40\,s latency barrier. After two additional experiments, it achieves a design with mean latency just above 30\,s—while maintaining low restart rates (about 20\%, mostly early in request lifetimes).

Next, the {\em Supervisor} encourages idea composition, recalling that \name had previously tested a {\em shortest-prompt-first} scheduler inspired by the classical shortest-job-first policy known to minimize mean completion time. \name {\em combines this idea} with HRA and implements the combined design. The result achieves a mean end-to-end latency under 23\,s—a 42.5\% improvement over the 40\,s baseline—with queueing delay reduced from 20\,s to just 3\,s. These improvements arise because the number of restarts reduces by 44\%. \name discovered this scheduler in only 20 simulations and in under two hours, compared to a human expert who required over 100 simulations and more than two weeks to reach similar insights.

A key advantage of \name’s agentic workflow is that it fosters {\em reasoning-driven exploration}. It formulates hypotheses, tests them empirically, and composes ideas, rather than relying solely on scalar feedback from benchmark scores. This reasoning process produces both higher performance and more interpretable designs than black-box optimization methods.

\paragraph{Continuous adaptation to changing workloads and application patterns.} A key motivation for automated decision-making is that optimal policies depend on factors that evolve over time.  Workload characteristics, hardware configurations, and application priorities can all shift, changing which routing policy performs best. For instance, the solution \name devised above is less effective in configurations with abundant KV-cache memory and short sequence lengths. Moreover, applications often optimize for different metrics: interactive services may emphasize TTFT and TPOT; background processing may prioritize overall throughput; and agentic or multi-stage applications may focus on end-to-end agent execution time. By running periodically and re-optimizing in response to observed conditions, \name can adapt its decisions to these evolving workloads and objectives.

\section{\name Agents}
\label{sec:design}
This section presents the design of \name’s agents. The current implementation comprises two primary agents:
\begin{enumerate}
    \item a {\em Researcher}, which proposes ideas, implements them, conducts experiments to generate metrics, and analyzes results; and 
    \item a {\em Supervisor}, which guides the Researcher by asking questions, providing feedback, and approving or suggesting revisions.
\end{enumerate}

We have taught the Researcher the general principles of systems research via its system prompt. The user prompt specifies the simulator and optimization problem to be solved.

To interact with simulator, the Researcher has access to shell commands and a cloned copy of the repository. As described in \S\ref{sec:motivation:sys}, it uses standard Unix commands (\eg \texttt{ls}, \texttt{grep}, \texttt{find}) to navigate the codebase, inspect directories, and identify relevant components—a process known as {\em agentic search}~\cite{anthropic2025effective}. The Researcher can create and execute scripts to analyze simulation outputs and performance metrics, enabling data-driven refinement of its hypotheses. Through these interactions, it conducts a human-like research process that combines experimentation and reasoning to discover new algorithms.

\name’s white-box, reasoning-driven workflow elevates exploration from the code level to the idea level. This makes its agentic research loop both more efficient (\Fig{overall}) and more interpretable (\S\ref{sub:overall_natural}) than prior black-box approaches such as FunSearch, AlphaEvolve, and EoH~\cite{FunSearch,AlphaEvolve,EoH}. Rather than relying solely on scalar performance scores, \name analyzes {\em why} a design succeeds or fails. It examines experimental evidence, forms hypotheses about root causes, and validates them through new experiments. As a result, its behavior more closely mirrors that of a human researcher, achieving both higher insight and robustness.

We first describe a single-context design in \S\ref{sub:single-context}, and then extend it to support multiple contexts in \S\ref{sub:multi-context}.

\subsection{Single-Context Glia (SCG)}
\label{sub:single-context}

Our initial design employs a single execution context shared by the two agents. Within this context, the Researcher may occasionally pursue unproductive directions, lose focus, or prematurely terminate exploration. When this occurs, the Supervisor intervenes to provide feedback and guidance: offering encouragement when the Researcher appears close to a promising idea, asking clarifying questions when potential directions are overlooked, and halting progress along clearly unproductive paths. The Supervisor also removes procedural obstacles, reminds the Researcher of overarching goals, and recalls previous findings. 

However, the Supervisor does not introduce new ideas or interfere directly with the Researcher’s reasoning. Unlike the Researcher, the Supervisor has no access to the codebase; it operates solely from the task description and from observing the Researcher’s outputs. In response to Supervisor's questions, the Researcher provides detailed rationales in a chain-of-thought style. This design intentionally mirrors the dynamics of a small human research team.

Maintaining a single execution context offers the benefit of a coherent, contiguous exploration history that is easily accessible to the Researcher.  However, it also introduces two limitations. First, current LLMs have finite context windows: once the limit is reached, the process must terminate, regardless of progress. Moreover, as the context grows, model attention becomes increasingly uneven across earlier content~\cite{liu-etal-2024-lost,anthropic2025effective}. Second, the single-context design scales poorly as there is no straightforward way to improve the resulting algorithms merely by increasing compute, budget, or iteration count. The next section introduces two extensions that mitigate these limitations.

\subsection{Multi-Context Glia (MCG)}
\label{sub:multi-context}
To overcome the limitations imposed by a single finite context window, we adopt a {\em best-of-$N$} sampling strategy, a common method for scaling test-time computation~\cite{snell2024scaling,cobbe2021training}. In this approach, we execute $N$ independent single-context instances of \name, each exploring the design space along a distinct trajectory, and select the best-performing algorithm discovered across all runs based on benchmark scores.

We present two variants of this strategy: \textbf{Sequential MCG} and \textbf{Parallel MCG-$N$}. In Sequential MCG, single-context \name runs execute one after another, allowing the process to stop as soon as a satisfactory result is achieved (up to a maximum number governed by cost). In Parallel MCG-$N$, $N$ independent runs proceed concurrently, with $N$ serving as a user-specified hyperparameter. For both variants, \name returns the best-performing design across all runs as the final output. A performance comparison between these modes is presented in \S\ref{sec:multi-context-ablation}.

Because the runtime and output quality of a single-context instance cannot be predicted in advance, the two modes expose different trade-offs.  Sequential MCG achieves faster early progress, as results can be assessed incrementally after each run, but its early performance varies depending on which trajectories complete first. 

Parallel MCG-$N$, in contrast, requires more peak compute but provides steadier improvement: evaluating multiple trajectories simultaneously reduces the likelihood that all runs perform poorly, leading to more consistent aggregate performance. In practice, Sequential MCG is preferable when compute resources are constrained, while Parallel MCG-$N$ offers lower variance and smoother convergence. Our experiments show that most of the benefit saturates by $N{=}4$, with larger values yielding diminishing returns. We discuss these differences further in \S\ref{sec:multi-context-ablation}.
\section{Evaluation}
\label{sec:eval}
We use OpenAI \texttt{o3}~\cite{openai_o3_o4_mini_system_card} as the underlying language model. In the following experiments, we specified a budget of \$30 per optimization.

We evaluate \name in \texttt{vidur}, a simulator for distributed LLM serving systems~\cite{agrawal2024vidur}. This section focuses on the request routing strategy among LLM replicas in distributed serving environments. We study how \name discovers novel routing algorithms that lower the mean request completion time (RT) across all requests.

We simulate an LLM inference workload (ShareGPT~\cite{sharegpt}) on four NVIDIA A10 GPUs running Llama-3-8B-Instruct. To emulate reasoning workloads with heavy-tailed sequence length distributions, we independently inflate 5\% of decode lengths and 5\% of prompt lengths by 10$\times$. The workload uses a query rate of 7.5 queries per second (QPS), with bursty interarrival times following a log-normal distribution ($\sigma=2$). Each LLM replica uses chunked prefill~\cite{sarathi} for batch scheduling, with a chunk size of 8192. We run each experiment with ten random seeds.

The primary evaluation metric is mean request completion time (RT), defined as the end-to-end latency to receive a response. This reflects user-perceived responsiveness in LLM serving and measures the effectiveness of scheduling policies under dynamic workloads. We also report 90\% bootstrapped confidence intervals.

\NewPara{Comparisons to other approaches.} We compare \name to three state-of-the-art frameworks that employ LLMs and evolutionary strategies for discovery: (i) Evolution of Heuristics (EoH)~\cite{EoH}, (ii) FunSearch~\cite{FunSearch}, and (iii) OpenEvolve~\cite{openevolve}. The parameter settings for these systems are shown in \Tab{params}. For FunSearch, we restrict each island to 20 algorithms.


Evolution of Heuristics~\cite{EoH} introduces a ``thought'' step into algorithm design and evolves candidate heuristics using five operators: crossover operators ($E_1$: generate diverse heuristics, $E_2$: recombine common ideas) and mutation operators ($M_1$: modify heuristics for improvement, $M_2$: adjust parameters within a heuristic, $M_3$: simplify by removing redundancies). These operators iteratively evolve heuristic candidates.

FunSearch~\cite{FunSearch} applies a best-shot optimization within an island model. In each generation, it selects top-performing code examples from the solution database and prompts the LLM to refine them, maintaining a bounded candidate island through iterative improvement.

OpenEvolve~\cite{openevolve}, an open-source implementation of AlphaEvolve~\cite{AlphaEvolve}, also employs island-based evolutionary search. It begins with an initial candidate program and a task-specific evaluation function to maximize a score. A prompt sampler generates inputs for the LLM, which produces new program variants. The system then applies evolutionary principles and migration across islands to improve solutions across generations.

\NewPara{Baseline routing comparisons.} We compare \name with three routing heuristics from the literature: Round-Robin, Least-Loaded Queue~(\textbf{LLQ}), and Least Outstanding Requests~(\textbf{LOR}). \textbf{Round-Robin} forwards requests cyclically across replicas. LLQ selects the replica with the fewest inflight requests, while LOR chooses the one with the fewest waiting requests (those without allocated GPU memory). We also compare against an expert-designed, workload-specialized algorithm developed over two weeks by a senior systems researcher with over 20 years of experience.

\subsection{Glia outperforms baselines}
\label{sub:overall_bar}

\begin{figure}
\begin{subfigure}[b]{\linewidth}
    \centering
    \includegraphics[width=0.9\linewidth]{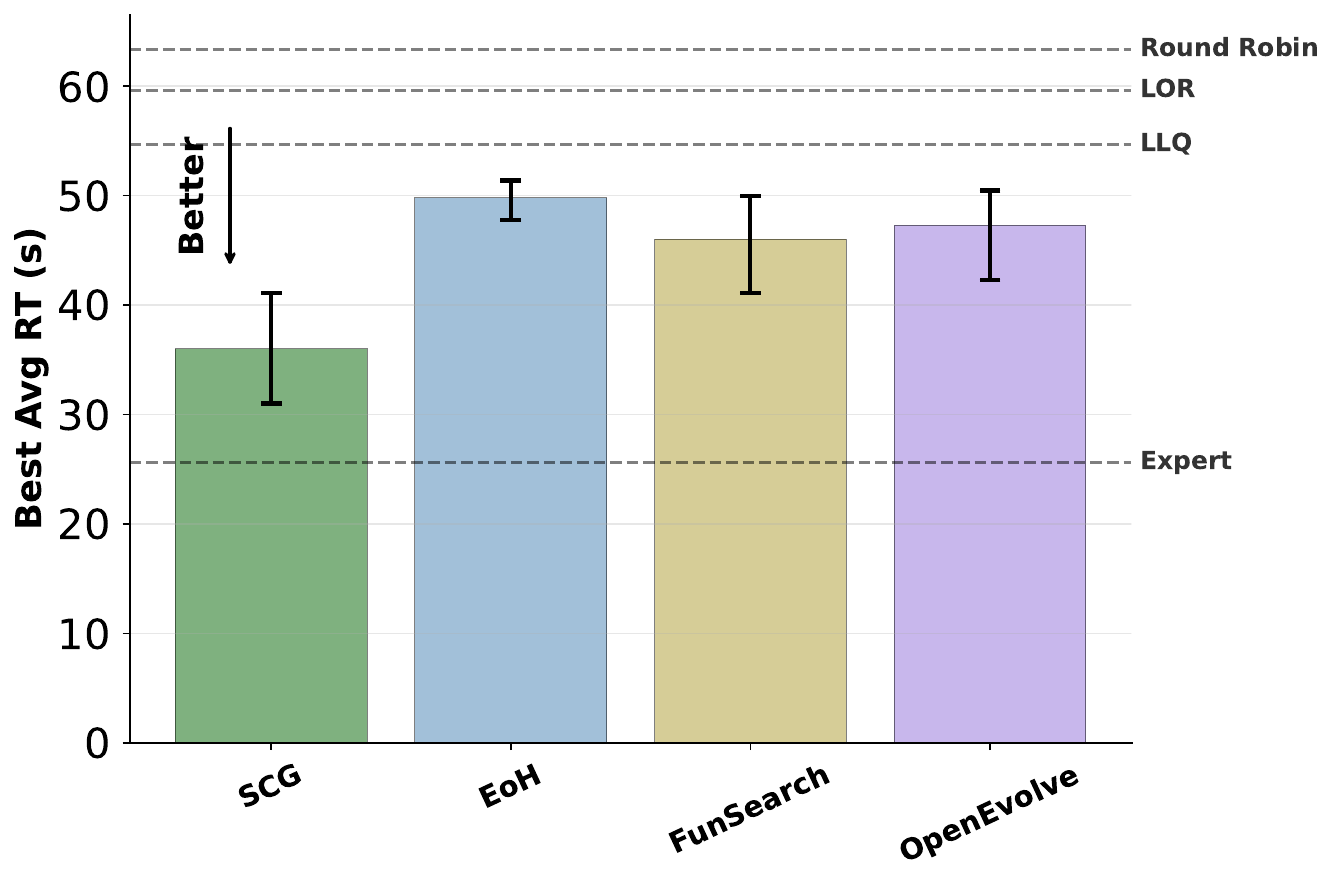}
    \caption{Single-Context \name (SCG) finds better algorithms compared to prior methods after a strict 15-simulation run budget. Lower response time (RT) is better. Error bars show the 90\% bootstrapping confidence intervals.}
    \label{fig:overall_bar_limited}
\end{subfigure}
\begin{subfigure}[b]{\linewidth}
    \centering
    \includegraphics[width=0.9\linewidth]{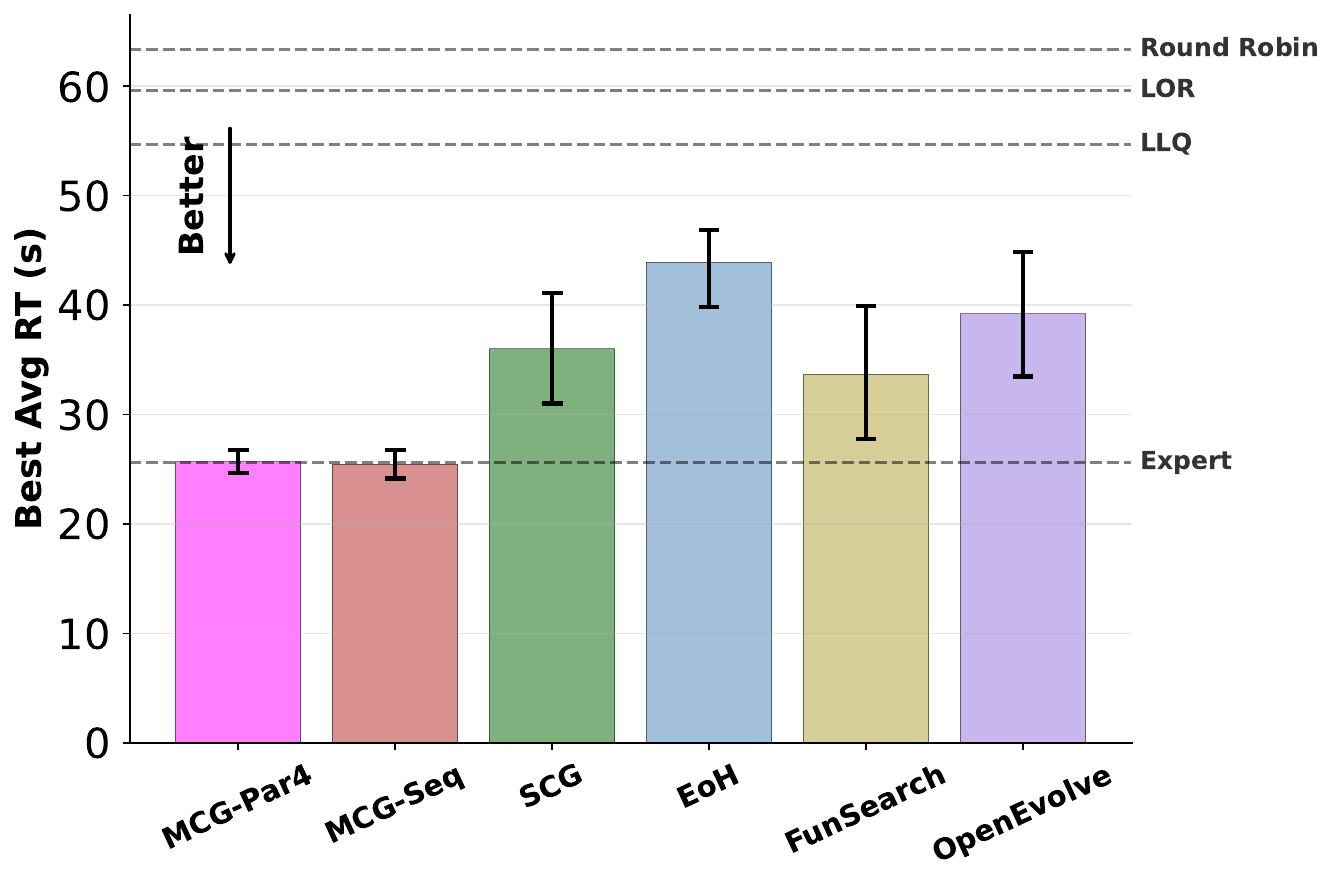}
    \caption{Comparison of the algorithms produced by \name with prior methods on a more relaxed 100-simulation budget (lower RT is better). Both versions of MCG \name (MCG-Par4 and MCG-Seq) find solutions that perform better than all the other methods. The error bars show the 90\% bootstrapping confidence intervals.}
    \label{fig:overall_bar}
\end{subfigure}
\end{figure}

\Fig{overall_bar_limited} compares the algorithms generated by Single-Context Glia (SCG) with routing heuristic baselines and prior algorithm design methods under a strict simulation budget of at most 15 runs. On average, SCG discovers algorithms that reduce mean response time (RT) by 1.3–1.4$\times$ compared to EoH, FunSearch, and OpenEvolve, and achieves the fastest discovery rate among them.

Next, we increase the simulation budget to 100 runs. \Fig{overall_bar} shows the best-performing algorithms produced by \name variants compared with the same baselines in this case. SCG performs similarly to the 15-simulation setting (\Fig{overall_bar_limited}). This stability stems from SCG’s stopping rule: it halts once the Researcher agent either completes its reasoning process or exhausts the context window, preventing further additions or revisions. In contrast, EoH, FunSearch, and OpenEvolve continue improving with additional simulations, attempting to use the extra resources. SCG cannot benefit from these available resources because it follows a single reasoning trajectory; because it runs out of context, it cannot extend its research process.

Multi-Context \name (MCG) addresses this limitation, as explained in \S\ref{sub:multi-context}. MCG samples multiple independent reasoning chains, either in parallel (MCG-Par4) or sequentially (MCG-Seq), and selects the best resulting algorithms. MCG-Par4 launches four independent SCG processes simultaneously to diversify the search, while MCG-Seq starts a new SCG after each previous run completes. Both MCG versions achieve the lowest average RT, outperforming SCG, traditional routing heuristics (Round-Robin, LLQ, LOR), and state-of-the-art LLM-based design frameworks (EoH, FunSearch, OpenEvolve). This multi-context scaling enables Glia to effectively utilize larger simulation budgets and continually improve solution quality.

MCG delivers substantial gains. On average its algorithms reduce average RT by \textbf{1.4$\times$} compared to SCG. It outperforms EoH, OpenEvolve, and FunSearch by \textbf{1.7$\times$}, \textbf{1.6$\times$}, and \textbf{1.3$\times$}, respectively.

\subsection{\name’s discoveries transfer to real systems} 

We implemented \name's routing strategy in Production Stack~\cite{production_stack}, an open-source request router for LLM inference using vLLM. Our testbed comprises 4 Nvidia A10 GPUs serving the Meta-Llama-3-8B-Instruct model. To evaluate, we measure \textbf{Slowdown}, defined as
\[
\text{Slowdown} = \frac{RT_{\text{system}}}{RT_{\text{ideal}}},
\]

where $RT_{\text{system}}$ is the request response time (RT) under the evaluated system, and $RT_{\text{ideal}}$ is the minimum possible RT in an ideal, load-free setting. A slowdown greater than $1$ indicates system overhead.

\begin{figure}
    \centering
    \includegraphics[width=0.9\linewidth]{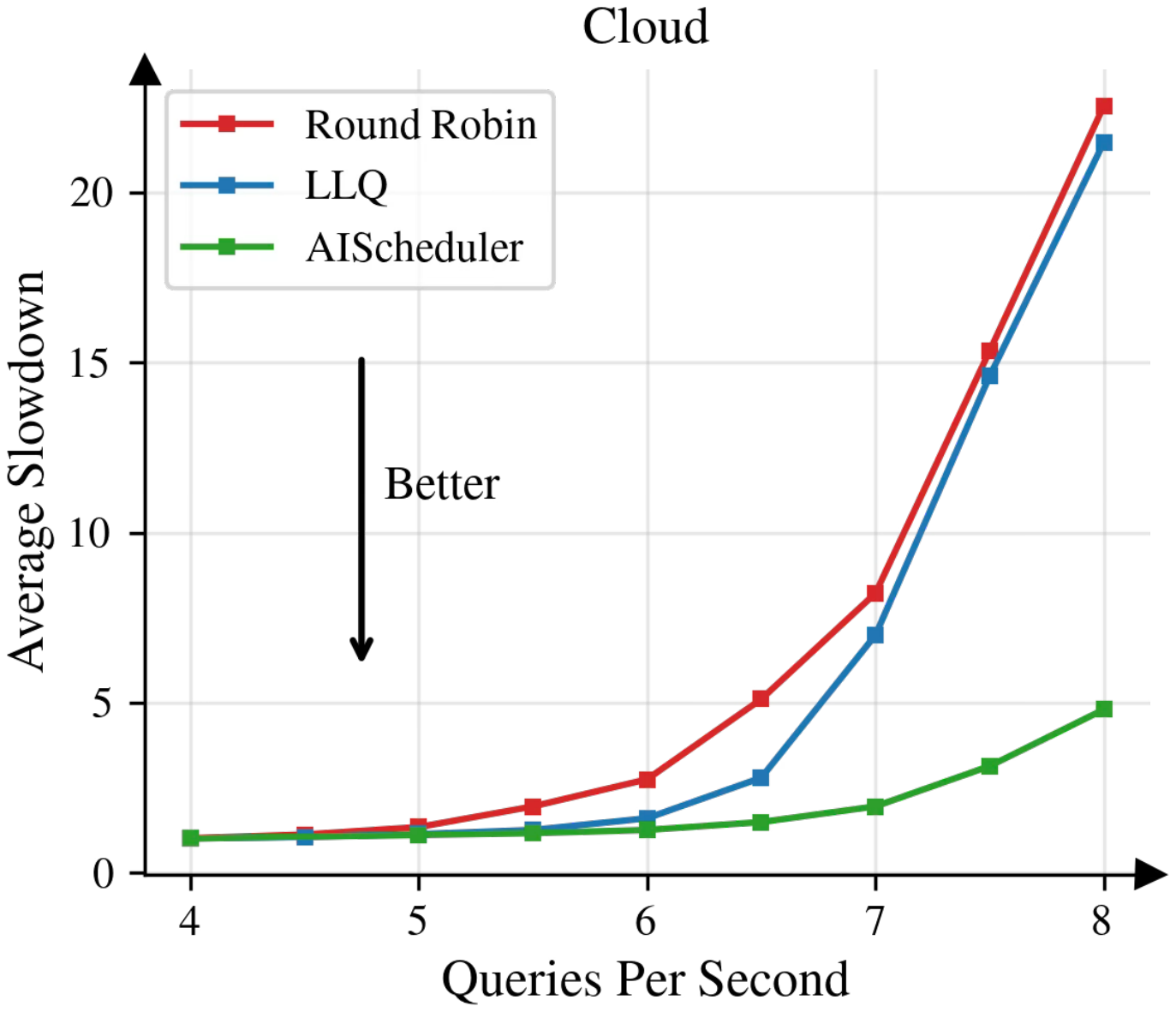}
    \caption{\name's discovered routing algorithm (AIScheduler in the figure) outperforms baselines in cloud experiments. The trends observed in the cloud experiments are similar to simulation though the numbers aren’t identical.}
    \label{fig:real}
\end{figure}

\Fig{real} shows, even under real-world workload variability and system uncertainty, \name’s discovered algorithm consistently outperforms all baselines. In the same load regime (QPS = 7.5), the router reduces slowdown by over 4.5$\times$ compared to LLQ, confirming Glia’s discoveries transfer effectively from simulation to a real system.

\subsection{Glia finds new algorithms across the stack}

We also applied \name to the vLLM {\em batch scheduler} and to the design of an {\em autoscaler} for the distributed vLLM cluster. 

Within the inference engine, whenever a GPU becomes idle, a scheduling algorithm selects which unfinished requests should proceed. This batch scheduler forms batches while respecting constraints such as available KV cache memory, maximum requests per batch, and token limits. In the experiments above, we used the Sarathi~\cite{sarathi} batch scheduler, which performs chunked prefill similarly to current vLLM implementations.

Using our optimized request router, we asked \name to improve the batch scheduler. It discovered that ordering requests by prefill length (rather than arrival time, as in vLLM and Sarathi) reduces end-to-end delay by an additional 25\%. The insight mirrors why Shortest-Remaining-Time-First outperforms First-Come-First-Served: prioritizing shorter prefills minimizes head-of-line blocking, reducing queueing delays for short prompts while adding negligible delay for longer ones, thereby lowering mean end-to-end latency.

At another layer of the orchestration stack, autoscaling adjusts the number of compute instances to meet latency targets while minimizing cost. To evaluate this, we implemented autoscaling in the vidur simulator and generated a long-running workload with temporal variability (QPS ranging from 7.5 to 22.5 in a slow sinusoidal pattern). We first implemented a baseline autoscaler, modeled after production systems, that scales based on per-instance decode throughput: it adds an instance when throughput exceeds a high threshold, removes the least busy instance when it falls below a low threshold, and includes a cooldown mechanism to prevent oscillation.

We then asked \name to design a more efficient autoscaler that minimizes compute cost while keeping the p95 slowdown below 5$\times$. \name proposed a proportional control loop that adjusts the number of instances based on inflight requests per instance. It then tuned the controller thresholds for this specific model and workload, finding an optimal configuration that minimizes cost while satisfying the latency constraint.

\begin{figure}
    \centering
    \includegraphics[width=0.9\linewidth]{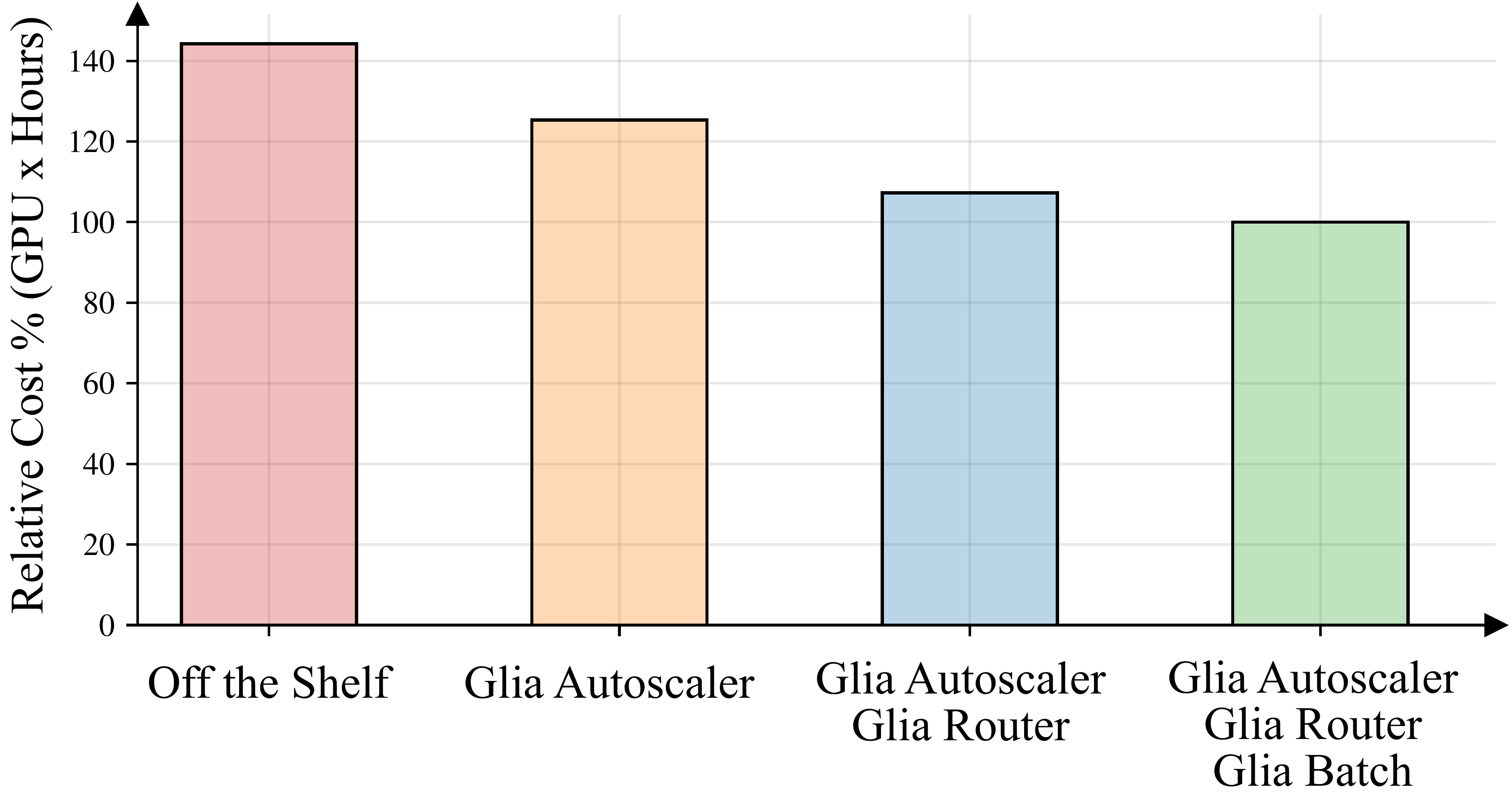}
    \caption{\name's GPU cost reductions as we progressively use it across the inference stack.}
    \label{fig:gpu_savings}
\end{figure}

\Cref{fig:gpu_savings} shows the total GPU$\times$hours saved when applying \name across different layers of the stack. The \name-discovered autoscaler alone reduces GPU cost by 13\% compared to an off-the-shelf autoscaler, while the full Glia-optimized stack (router, batch scheduler, and autoscaler) cuts total GPU$\times$hours by $40\%$ for this variable workload, compared to standard serving systems (vLLM batch scheduler, LLQ router, and throughput-based autoscalers).

\subsection{\name finds novel, interpretable algorithms}
\label{sub:overall_natural}
\Fig{single-context-glia} shows a representative algorithm discovered by \name. This algorithm is principled and, to the best of our knowledge, introduces a new concept in this domain: a {\em Head-Room Admission} (HRA) router that reserves headroom to accommodate unknown decode growths. The router follows an admission-control approach—if a replica lacks sufficient KV-cache memory (after accounting for headroom), the request is queued and dispatched only when resources may have become available. It combines this idea with a shortest-prefill-first policy, approximating shortest-job-first scheduling, a strategy known to minimize mean latency~\cite{Silberschatz:2018:OSC:3219243}. By maintaining a small KV-cache headroom on each replica at admission time, the router effectively prevents vLLM from running out of memory, thereby avoiding request restarts and wasted computation.

\name’s discovered algorithms are natural and intuitive, often reflecting the kind of principled reasoning a human designer might employ. In contrast, algorithms produced by prior evolutionary frameworks such as EoH, FunSearch, and OpenEvolve are typically more complex and less interpretable. For instance, the program generated by FunSearch (\Fig{funsearch-code}) includes numerous hyperparameters, conditional branches, and opaque thresholds, making it difficult to identify which components actually drive performance. Other evolutionary baselines show similar traits, yielding heuristics that are cumbersome to interpret and challenging to analyze, refine, or reason about.

\subsection{Glia can continuously adapt to changes}
\label{sec:eval_adaptation}

We next evaluate Glia for router optimization under a different \emph{i)}~workload, \emph{ii)}~hardware, and \emph{iii)}~optimization settings.
All experiments in this section are conducted within the simulator.
The Researcher agent uses \texttt{GPT5} as its underlying model.
\begin{itemize}
\item \textbf{Workload:} Prefill-heavy, with a prompt-to-decode ratio of about $70$. The workload generator operates in a closed loop, maintaining 200 concurrent requests in the system—analogous to a chat application with 200 users waiting for LLM responses before sending their next messages.
\item \textbf{Model and Hardware:} \texttt{Llama-3.3-70B-Instruct-FP8-dynamic} running on 8 Nvidia H100 GPUs.
\item \textbf{Objective:} Constrained optimization—maximize request throughput (QPS) while keeping P90 TTFT below 1500~ms.
\end{itemize}

In this new setting, none of the routing heuristics—Round-Robin, LLQ, or LOR—satisfy the TTFT constraint.
More importantly, the human-expert-designed heuristic (\S\ref{sub:overall_bar}) not only violates the constraint but also yields a significantly higher TTFT than these baseline routing heuristics.

This reveals a key challenge: the expert heuristic’s strong performance in \S\ref{sub:overall_bar} relied on workload specialization.

Without automated adaptation, a human expert would need to reanalyze each new workload and redesign the routing strategy, an expensive and time-consuming process.

\name overcomes this challenge by automatically adapting to new operating conditions.
Under the new workload, \name discovers a routing algorithm that meets the TTFT constraint in all ten trials.
Moreover, its achieved QPS exceeds that of the baseline heuristics, even though those heuristics fail to meet the TTFT constraint.
This example demonstrates \name’s versatility and ability to perform constrained optimization, ensuring SLO compliance without human intervention

\begin{figure}
    \centering
    \includegraphics[width=0.9\linewidth]{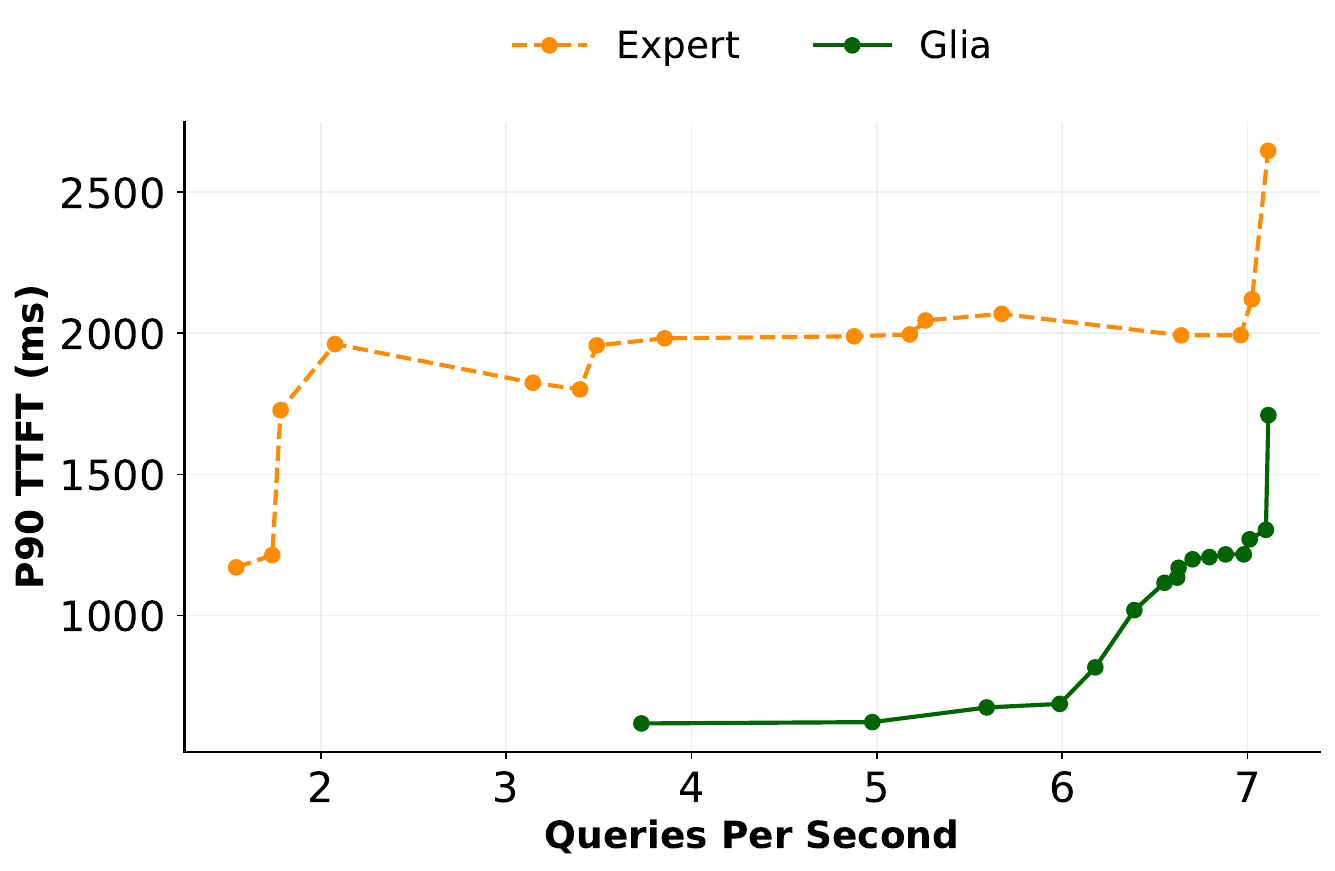}
    \caption{Trade-off between tail (P90) Time to First Token (TTFT) and request throughput for the expert-designed algorithm and \name-designed algorithm. The expert algorithm was tailored to a different problem setup, and struggles in this prefill-heavy workload.}
    \label{fig:problem_adaptation}
\end{figure}

To further analyze the \name-discovered routing algorithm, we vary the number of inflight requests and examine the trade-off between tail TTFT and request throughput (QPS) for both the expert-designed (\S\ref{sub:overall_bar}) and \name-designed algorithms in \Fig{problem_adaptation}.
As shown in the figure, the expert algorithm fails to maintain acceptable TTFT as QPS increases, satisfying the SLO only up to 1.5 QPS.
In contrast, the \name-discovered algorithm sustains compliant TTFT up to 7 QPS—a 4.6$\times$ improvement over the expert algorithm.

\subsection{Ablation experiments}
\label{sec:multi-context-ablation}

We conduct ablations to understand how \name's design choices affect performance. Appendix~\ref{sec:multi-context-ablation-details} provides the full results and figures.

\paragraph{\name finds good algorithms quickly.}
Single-Context Glia (SCG) achieves steep early gains, driven by coherent and continuous white-box reasoning within a single context. In our experiments, SCG discovers strong algorithms using fewer simulations than prior algorithm design methods.

However, SCG’s scalability is limited. Because the researcher can conclude the search regardless of available simulations, additional resources cannot easily improve performance. We address this limitation with Multi-Context Glia (MCG) approaches, which run multiple independent SCG processes and select the best result. As shown in Appendix~\ref{sec:multi-context-ablation-details}, both parallel and sequential MCG continue improving beyond SCG and achieve lower RT than prior methods.

\paragraph{Supervisor impact.}
We also observe that Supervisor interventions can help when the Researcher approaches a performance plateau. In such cases, the Supervisor encourages revisiting earlier ideas, checking overlooked combinations, and composing mechanisms that had previously been evaluated in isolation. Appendix~\ref{sec:multi-context-ablation-details} includes an example intervention. We further quantify this effect by ablating the Supervisor agent: across 10 runs, removing the Supervisor leads to worse performance. The runs without the Supervisor also tend to plateau earlier, consistent with the Supervisor encouraging additional experiments before concluding.

\paragraph{Multi-context Glia ablation.}
Finally, we compare Multi-Context variants. $N$-way Parallel Glia runs $N$ independent instances of SCG in parallel and reports the best-performing algorithm among them. With more computational resources, more instances can be launched simultaneously, improving robustness to unproductive trajectories. In our experiments, larger $N$ yields diminishing returns beyond modest values. In contrast, Sequential \name introduces no additional parameters and is simpler to use, though it can take longer to match the best parallel configuration. Full comparisons are in Appendix~\ref{sec:multi-context-ablation-details}.

\section{Conclusion}
\label{sec:conclusion}
We are progressing toward our primary goal: developing Glia into an AI capable of PhD-level systems design and optimization for real-world problems. While this paper's focus is on AI inference (covering both large language models and traditional AI workloads), our broader vision is a general-purpose system architect that autonomously enhances the performance, efficiency, and adaptability of emerging computing systems.

\name demonstrates the promise of AI-driven infrastructure optimization, yet significant open questions remain. Achieving fully self-managing computing infrastructures will require progress across several key dimensions:
\begin{enumerate}
\item \textbf{Robustness and safety}: Automated systems must remain stable under adversarial or unexpected conditions. AI-based approaches must ensure robustness to workload shifts, failures, and anomalies.
\item \textbf{Human-AI collaboration}: Glia’s explainability is a core strength, and ongoing work aims to expand how system architects engage with AI-generated insights. Richer visualization, interactive debugging, and co-design workflows could further enhance productivity and trust.

\item \textbf{Generalization across systems}: Although current results focus on AI inference stacks, our approach holds promise for optimizing a broad range of complex computer systems. 
\item \textbf{Abstractions and architectural discovery}: Designing a new abstraction such as an API, transport semantic, or queueing primitive is qualitatively different from optimizing a heuristic policy. Abstraction design requires reasoning about what to expose or hide, composability, failure modes, and long-term maintainability. These qualities are less measurable and typically evaluated through argumentation, analysis, and targeted experiments. Whether AI can propose abstractions that are both useful and stable is an open question.

\end{enumerate}

\newpage
\bibliographystyle{ACM-Reference-Format}


\newpage
\clearpage
\appendix
\section{Appendix}

\begin{table}[t]
\centering
\resizebox{0.9\columnwidth}{!}{
\begin{tabular}{lll}
\toprule
\textbf{System} & \textbf{Parameter Name} & \textbf{Value} \\
\midrule
\multirow{1}{*}{EoH} 
  & Initial population size (N) & 20 \\
\midrule
\multirow{1}{*}{FunSearch} 
  & \# Prompt programs (k) & 3 \\
\midrule
\multirow{5}{*}{OpenEvolve} 
  & Number of islands & 6 \\
  & Exploration probability & 0.6 \\
  & Exploitation probability & 0.4 \\
  & Initial program & LLQ router \\
\bottomrule
\end{tabular}
}
\caption{Hyperparameters for the other approaches used in our evaluation.}
\label{tab:params}
\end{table}
\subsection{Ablation experiments details}
\label{sec:multi-context-ablation-details}

\begin{figure}
    \centering
    \includegraphics[width=\linewidth]{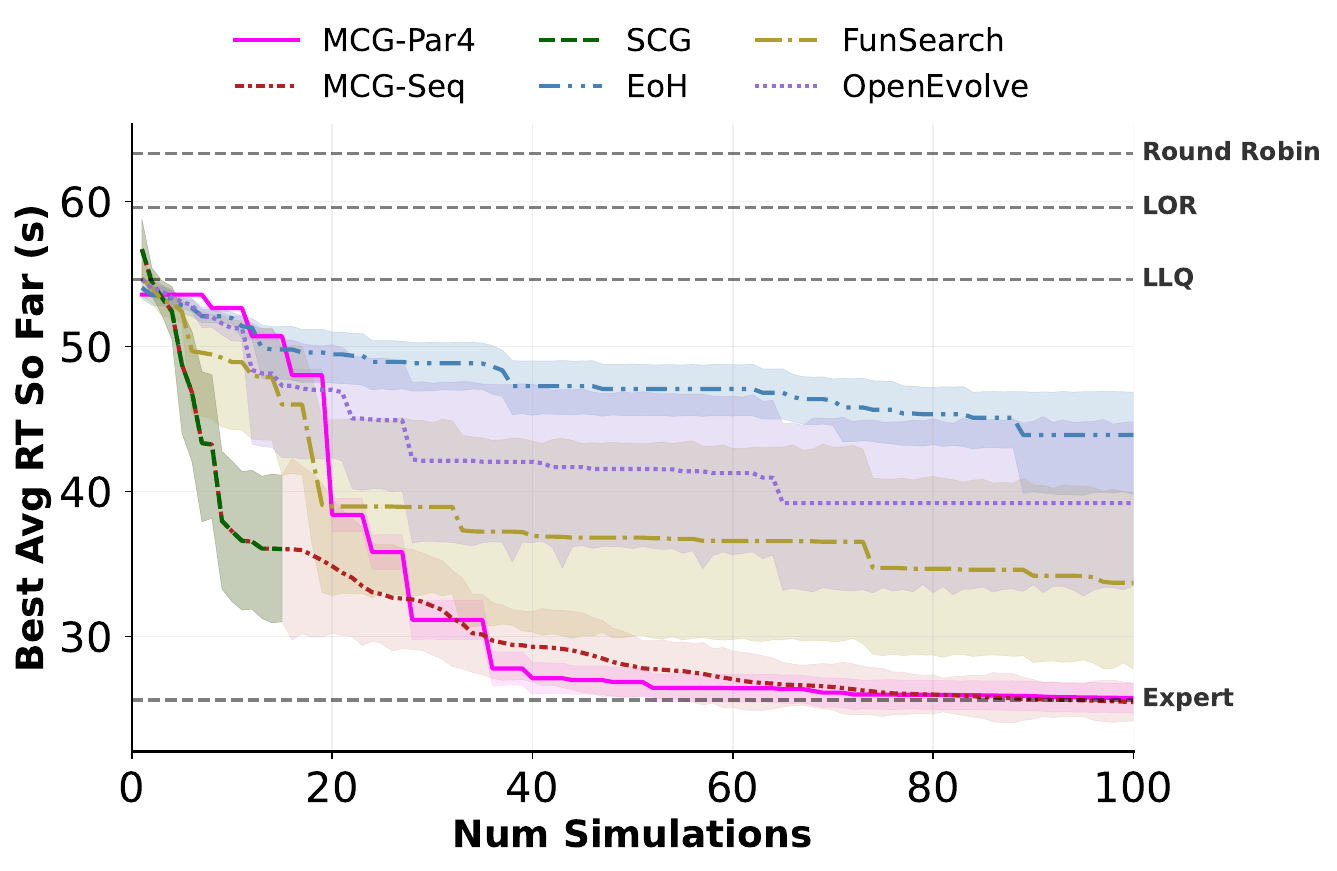}
    \caption{Comparison of \name variants with baselines and prior methods (lower is better). SCG has the steepest early gains thanks to coherent and continuous white-box reasoning. The two variants of MCG---4-way parallel (MCG-Par4) and sequential (MCG-Seq)---extend the gains and outperform other methods by finding better algorithms more quickly. Shades show 90\% confidence intervals.}
    \label{fig:overall}
\end{figure}

\paragraph{\name finds good algorithms quickly.} \Fig{overall} compares Glia’s progress across simulations with prior algorithm design methods (lower is better). Glia consistently discovers stronger algorithms using far fewer simulations. Single-Context Glia (SCG) achieves the steepest early gains, driven by white-box reasoning and focused, continuous refinement within a single context.

However, SCG’s scalability is limited. Because the researcher can conclude the search regardless of available simulations (hence the shorter SCG curve), additional resources cannot easily improve performance. We address this limitation with Multi-Context Glia (MCG) approaches. As shown in \Fig{overall}, both 4-way Parallel Glia (MCG-Par4) and Sequential Glia (MCG-Seq) overcome SCG’s constraint and continue to improve. On average, MCG methods discover better-performing algorithms faster than EoH, FunSearch, and OpenEvolve, achieving lower RT than all other methods.

\paragraph{Supervisor impact.} To illustrate the Supervisor’s impact, we present an example intervention. In one run, early exploration produced two ideas: (i) combining backlog-aware selection with heavy-request isolation, and (ii) bounding the one-pass queue scan to prevent large requests from starving. Instead, the Researcher pursued adaptive headroom tuning and heavy-lane gating, which reduced latency from 55~s to 28~s but still missed the 22~s target. Believing further progress was unlikely, the Researcher considered stopping. At this point, the supervisor intervened:

\begin{tcolorbox}[
  colback=blue!5!white,    
  colframe=blue!75!black,  
  rounded corners          
]
\small
You’re close, but not stuck yet. A couple of quick checkpoints:  

\begin{enumerate}
    \item Recall your finding that queueing delay still dominates whenever memory is oversubscribed.
    \begin{itemize}
        \item Does the dynamic-headroom variant fully leverage backlog information when deciding headroom, or is it applied replica-independently?
        \item You previously tested backlog-aware selection and heavy-request isolation separately; have you measured them together under the same admission-control knob?
    \end{itemize}

    \item Your one-pass queue scan leaves large, unsafe requests parked. Verify whether those accumulate and starve; if so, a bounded re-ordering window (still SPF) might smooth the tail without extra restarts.
\end{enumerate}

If you haven’t exhausted these combinations, worth a final sweep before concluding.
\end{tcolorbox}

The Researcher then proceeds to revisit these earlier ideas, combining backlog-aware selection with heavy-request isolation and exploring bounded re-ordering, which ultimately advanced the router beyond the plateaued performance.

\begin{figure}[t]
  \centering
  \begin{subfigure}[b]{\linewidth}
    \centering
    \includegraphics[width=\linewidth]{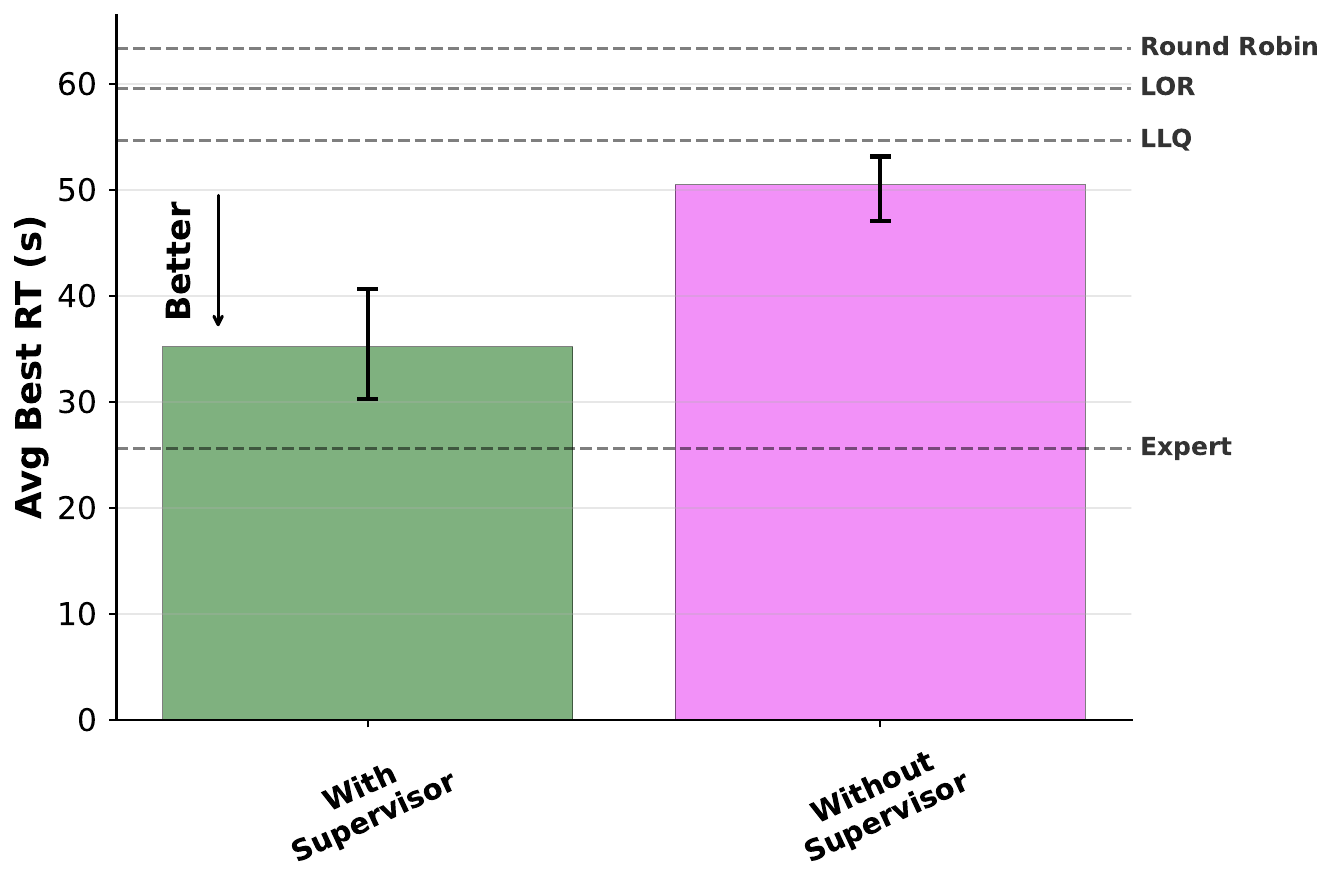}
    \caption{With the Supervisor, SCG finds better solutions (lower RT is better).}    \label{fig:supervisor_bar}
  \end{subfigure}
  \begin{subfigure}[b]{\linewidth}
    \centering
    \includegraphics[width=\linewidth]{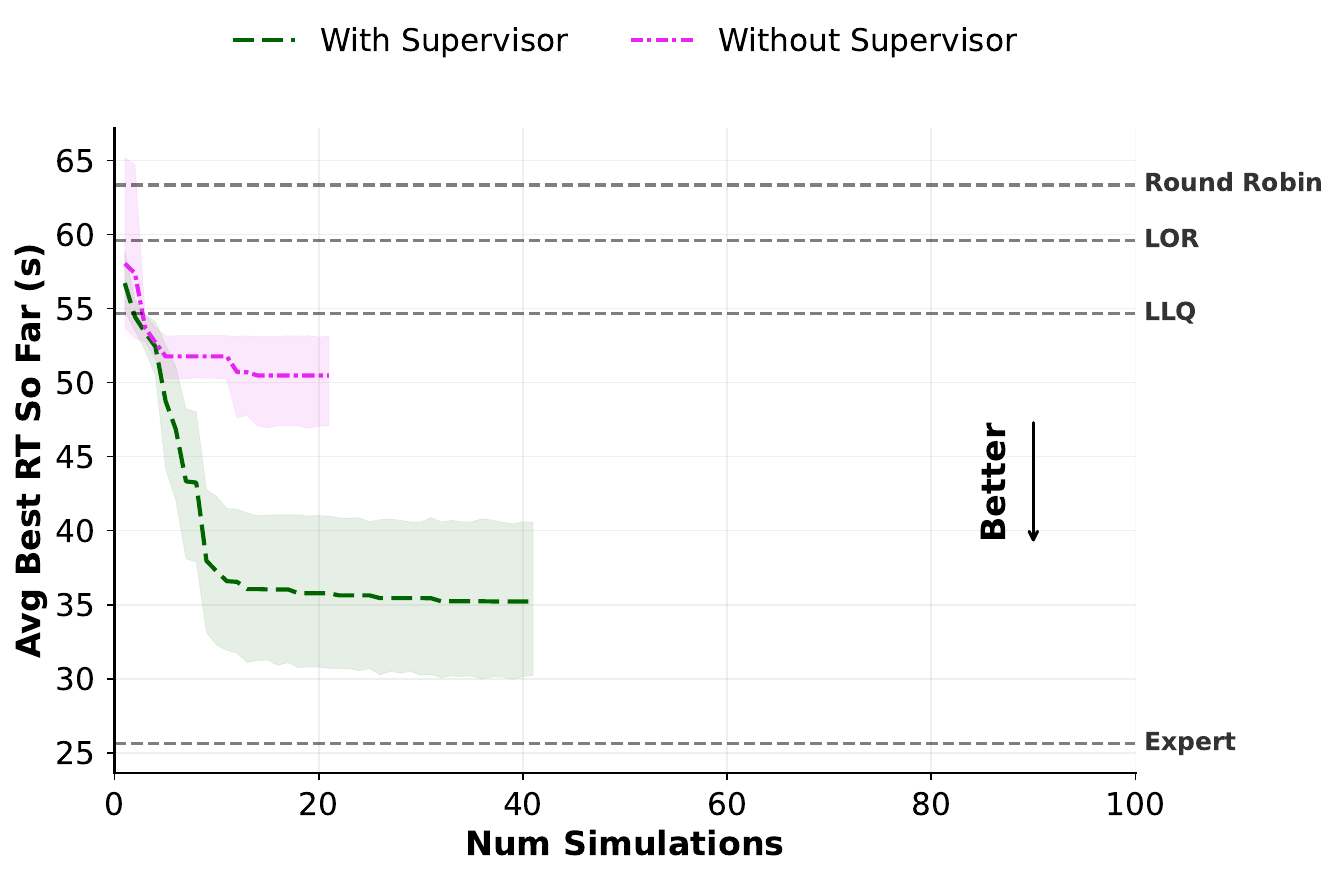}
    \caption{The Supervisor encourages SCG to continue improving and runs more simulations; without it, progress plateaus earlier.}    \label{fig:supervisor_progress}
  \end{subfigure}
  \caption{\textbf{Ablation of the Supervisor agent.} We run Single-Context Glia (SCG) with and without the Supervisor for 10 runs. (a) Best average RT achieved (error bars show 90\% bootstrapped confidence intervals). (b) Best-so-far RT versus number of simulations; shaded regions show variability across runs. Horizontal dashed lines indicate baseline heuristics and the expert.}
  \label{fig:supervisor_ablation}
\end{figure}

We further quantify the Supervisor’s impact by ablating it entirely. We run SCG for 10 runs with the Supervisor removed, keeping the rest of the setup unchanged. \Fig{supervisor_ablation} shows that removing the Supervisor degrades the best solution quality: with the Supervisor, SCG achieves substantially lower best average RT than without it (\Fig{supervisor_bar}). 

\Fig{supervisor_progress} suggests a second effect: without the Supervisor, the Researcher tends to plateau earlier and the run terminates after fewer simulations, whereas Supervisor guidance sustains exploration and encourages the researcher agent for additional experiments. 

\begin{figure}
    \centering
    \includegraphics[width=0.9\linewidth]{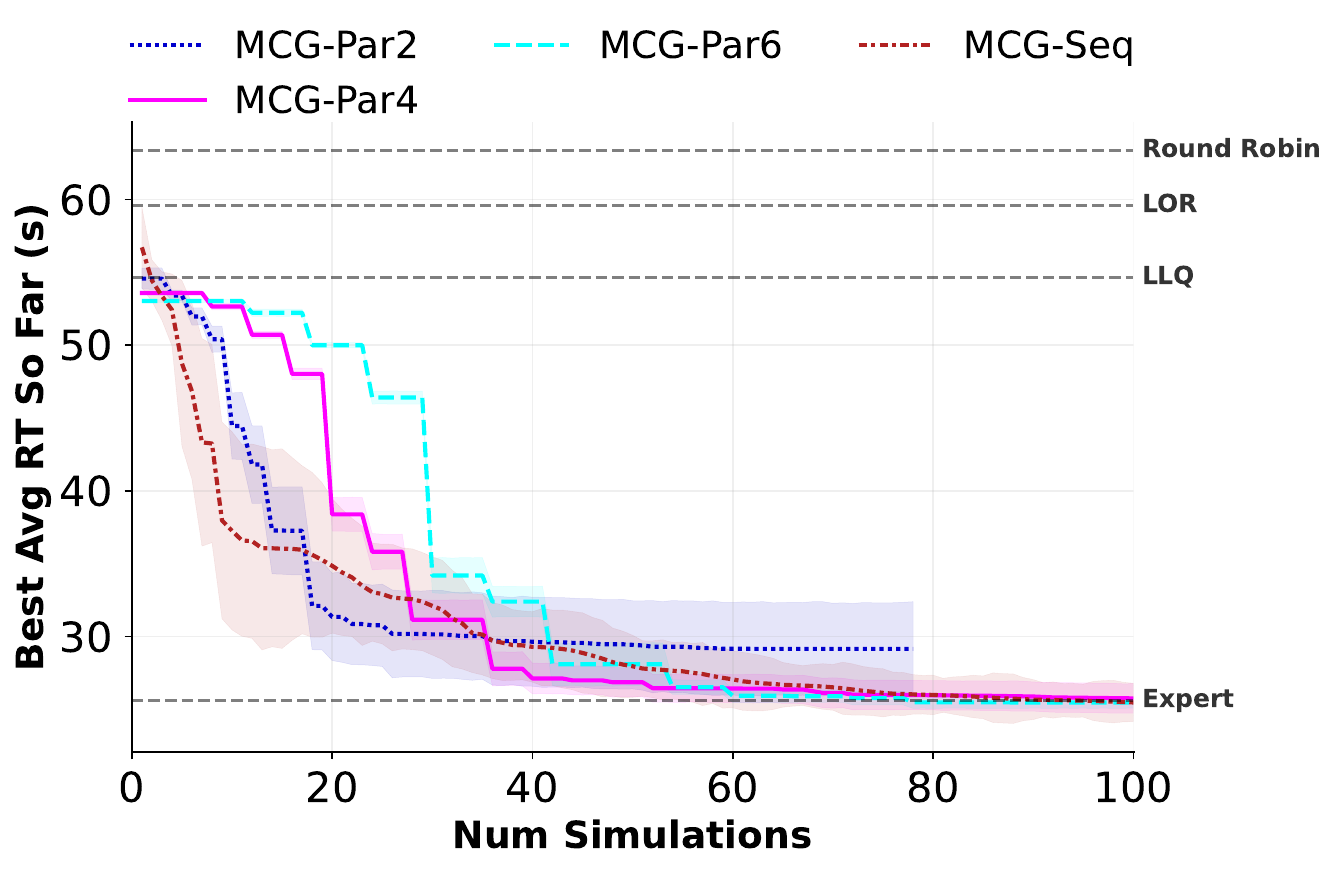}
    \caption{Comparing Glia variants (lower is better).}
    \label{fig:best_of_n}
\end{figure}

\paragraph{Multi-context Glia ablation.}
\Fig{best_of_n} compares Glia's Multi-Context variants.
$N$-way Parallel Glia runs $N$ independent instances of Single-Context Glia in parallel and reports the best-performing algorithm among them. Its key advantage is scalability: with more computational resources, more instances can be launched simultaneously. For example, 4-way Parallel Glia outperforms 2-way because the likelihood that all agents “go astray” decreases as the number of parallel runs increases. This design also mitigates early exploration risk, since at least one agent is likely to discover a strong algorithm quickly. However, it introduces a new hyperparameter, $N$, which must be chosen; there is no obvious optimal value, though in our experiments $N=4$ achieved a good balance between performance and resource cost. Larger values (e.g., $N=6$ in \Fig{best_of_n}) yield diminishing returns.

In contrast, Sequential \name requires no additional parameters and is simpler to use. It initially reduces error more sharply than $N$-way Parallel (within the first 35 simulations) but later slows down, taking longer to match the performance of 4-way Parallel Glia.

\onecolumn

\captionsetup{type=figure}
\captionof{figure}{Python code for the Head-Room Allocator (HRA) request routing algorithm discovered by \name.}
\label{fig:single-context-glia}
\begin{lstlisting}[style=pythonstyle, inputencoding=utf8]
"""Head-Room Admission (HRA) global scheduler.

This scheduler mitigates vLLM pre-emptions by keeping a small KV-cache
head-room on every replica *at admission time*.  For each incoming request we
pessimistically reserve additional blocks to account for the (unknown) decode
phase and admit the request only if the target replica would still retain the
configured safety margin.

Empirical defaults (good for the ShareGPT-style workload used in Vidur's
benchmarks):

    DECODE_TO_PREFILL_RATIO  = 0.6   # avg decode/prompt tokens
    SAFETY_FRACTION          = 0.03  # keep last 3 % blocks free

These values reduce average end-to-end latency by ~40 % compared to LLQ while
maintaining >95 % GPU utilisation.
"""

from math import ceil
from typing import List, Tuple

from vidur.entities import Request
from vidur.scheduler.global_scheduler.base_global_scheduler import BaseGlobalScheduler

# ---------------------------------------------------------------------------
# Tunable constants (change if workload characteristics differ significantly)
# ---------------------------------------------------------------------------

DECODE_TO_PREFILL_RATIO: float = 0.6   # pessimistic decode growth factor
SAFETY_FRACTION: float = 0.03          # minimum fraction of blocks kept free


class AIGlobalScheduler(BaseGlobalScheduler):
    """Memory-aware global scheduler with fixed head-room admission control."""

    # pylint: disable=protected-access

    def schedule(self) -> List[Tuple[int, Request]]:
        # Always serve the *shortest* prompt next (SJF) to minimise mean latency.
        self._request_queue.sort(key=lambda r: (r.num_prefill_tokens, r.arrived_at))

        if not self._request_queue:
            return []

        # Cluster-wide, all replicas share the same memory configuration.
        any_scheduler = next(iter(self._replica_schedulers.values()))
        block_size = any_scheduler._config.block_size
        max_blocks = any_scheduler._config.num_blocks
        min_free_blocks = int(max_blocks * SAFETY_FRACTION)

        # Snapshot per-replica state and keep optimistic updates locally so that
        # multiple placements within one call are consistent.
        allocated_blocks = {
            rid: rs.num_allocated_blocks for rid, rs in self._replica_schedulers.items()
        }
        pending_reserved_blocks = {
            rid: ceil(
                sum(r.num_prefill_tokens * (1 + DECODE_TO_PREFILL_RATIO) for r in rs._request_queue)
                / block_size
            )
            for rid, rs in self._replica_schedulers.items()
        }
        queue_lengths = {
            rid: rs.num_pending_requests + rs.num_active_requests for rid, rs in self._replica_schedulers.items()
        }

        request_mapping: List[Tuple[int, Request]] = []

        idx = 0
        # Traverse requests in order; if the head request cannot be admitted we
        # stop to preserve FIFO fairness (new arrivals behind it must wait).
        while idx < len(self._request_queue):
            req = self._request_queue[idx]

            req_blocks = ceil(
                req.num_prefill_tokens * (1 + DECODE_TO_PREFILL_RATIO) / block_size
            )

            admissible = []
            for rid in self._replica_schedulers.keys():
                projected_usage = allocated_blocks[rid] + pending_reserved_blocks[rid] + req_blocks
                free_after = max_blocks - projected_usage
                if free_after >= min_free_blocks:
                    admissible.append(rid)

            if not admissible:
                break  # cannot place the oldest waiting request right now

            # Choose replica with lowest projected usage; tie-break by queue length.
            target_rid = min(
                admissible,
                key=lambda rid: (allocated_blocks[rid] + pending_reserved_blocks[rid], queue_lengths[rid]),
            )

            # Commit placement and optimistically update state for subsequent decisions.
            request_mapping.append((target_rid, req))
            self._request_queue.pop(idx)  # do *not* increment idx

            pending_reserved_blocks[target_rid] += req_blocks
            queue_lengths[target_rid] += 1

        return request_mapping
\end{lstlisting}

\captionsetup{type=figure}
\captionof{figure}{Code generated by FunSearch.}
\label{fig:funsearch-code}
\begin{lstlisting}[style=pythonstyle, inputencoding=utf8]
class CustomGlobalScheduler(BaseGlobalScheduler):  # type: ignore[name-defined]
    """Latency-oriented, eviction-aware global scheduler.

    Key features
    -------------
    1.  Decode length prediction per *prefill* bucket (small / mid / large)
        with an online exponential moving average; gives markedly better
        memory-footprint forecasts than a single global estimate.

    2.  Looks ahead and keeps a projection of every replica's future state
        (memory blocks, remaining pre-fill backlog, queue length).  The
        projection is updated greedily after each assignment so later
        decisions use a consistent view.

    3.  Request priority is *rescue-first SJF*: previously evicted jobs first
        (to avoid starvation / wasted work), then smaller **total** expected
        tokens, finally FIFO.

    4.  Replica selection minimises a composite cost of projected memory
        utilisation (quadratic), outstanding pre-fill backlog, queue length
        and the *instant* block deficit for the pre-fill of the candidate
        request.  Jobs with restarts receive a multiplicative cost discount.

    5.  Admission control: a new request is dispatched only if the projected
        utilisation stays below a configurable soft limit.  The limit is
        relaxed slightly for restarted jobs so they can finish.
    """

    # ------------- tunables -------------
    _BUCKET_BOUNDS = (128, 512)       # <128 small, 128-512 mid, >512 large
    _EMA_ALPHA = 0.10                 # smoothing for bucketed averages
    _INIT_DECODE_EST = 96.0           # bootstrap decode len (tokens)
    _MIN_DECODE = 32.0
    _MAX_DECODE = 1024.0

    _SOFT_UTIL_CAP = 1.03             # ordinary requests must stay under this
    _SOFT_UTIL_CAP_RESTART = 1.10     # restarted jobs may exceed slightly

    # cost weights (should sum ~1)
    _W_UTIL = 0.55                    # projected utilisation^2
    _W_BACKLOG = 0.25                 # remaining pre-fill backlog fraction
    _W_QUEUE = 0.10                   # queue length fairness
    _W_DEFICIT = 0.10                 # instantaneous block deficit

    _OVER_CAP_PEN = 12.0              # extra when util>1
    _RESTART_DISCOUNT = 0.6           # multiplicative cost discount per restart

    # ------------------------------------

    def __init__(self, *args: Any, **kwargs: Any):  # type: ignore[override]
        super().__init__(*args, **kwargs)

        # bucketed decode length EWMA statistics
        # structure: (count , avg)
        self._bucket_avg: List[float] = [self._INIT_DECODE_EST] * 3
        self._bucket_cnt: List[int] = [0, 0, 0]

        # global fallback EWMA
        self._global_avg: float = self._INIT_DECODE_EST

        # snapshot of visible requests from previous tick (id -> (pf, processed))
        self._prev_snapshot: Dict[int, Tuple[int, int]] = {}

    # --------------------- helper: bucket index ---------------------
    @classmethod
    def _bucket_idx(cls, prefill: int) -> int:
        if prefill < cls._BUCKET_BOUNDS[0]:
            return 0
        if prefill < cls._BUCKET_BOUNDS[1]:
            return 1
        return 2

    # --------------------- statistics maintenance ------------------
    def _update_decode_statistics(self) -> None:
        """Detect completed requests and update bucket/global decode EWMAs."""
        current: Dict[int, Tuple[int, int]] = {}

        # helper to insert into current snapshot quickly
        def _collect(req):
            current[id(req)] = (req.num_prefill_tokens, req.num_processed_tokens)

        for req in self._request_queue:
            _collect(req)
        for rep in self._replica_schedulers.values():
            for rq in rep.pending_queue:
                _collect(rq)
            for rq in rep.active_queue:
                _collect(rq)

        # detect finished requests
        finished_ids = set(self._prev_snapshot.keys()) - set(current.keys())
        for rid in finished_ids:
            pf_tokens, processed = self._prev_snapshot[rid]
            decode_tokens = max(0, processed - pf_tokens)
            if decode_tokens <= 0:
                continue

            # update bucket stats
            bidx = self._bucket_idx(pf_tokens)
            old_avg = self._bucket_avg[bidx]
            new_avg = (1.0 - self._EMA_ALPHA) * old_avg + self._EMA_ALPHA * decode_tokens
            self._bucket_avg[bidx] = min(max(new_avg, self._MIN_DECODE), self._MAX_DECODE)
            if self._bucket_cnt[bidx] < 1e9:  # avoid overflow
                self._bucket_cnt[bidx] += 1

            # update global average
            g_new = (1.0 - self._EMA_ALPHA) * self._global_avg + self._EMA_ALPHA * decode_tokens
            self._global_avg = min(max(g_new, self._MIN_DECODE), self._MAX_DECODE)

        self._prev_snapshot = current

    # -------------------- decode prediction ------------------------
    def _predict_decode(self, prefill_tokens: int) -> float:
        bidx = self._bucket_idx(prefill_tokens)
        if self._bucket_cnt[bidx] >= 10:  # need some data for bucket-specific
            return self._bucket_avg[bidx]
        return self._global_avg

    # -------------------- utility functions ------------------------
    @staticmethod
    def _ceil_div(a: float, b: int) -> int:
        return int(math.ceil(a / b))

    # ---------------------------- main ------------------------------
    def schedule(self) -> List[Tuple[int, 'Request']]:  # type: ignore[name-defined]
        # housekeeping
        self._update_decode_statistics()
        if not self._request_queue:
            return []

        replicas: Dict[int, 'ReplicaScheduler'] = self._replica_schedulers  # type: ignore[name-defined]
        num_repls: int = max(1, self._num_replicas)

        # ---------- priority sort for global queue ----------
        def _priority(req: 'Request') -> Tuple[int, float, float]:  # type: ignore[name-defined]
            predicted_total = req.num_prefill_tokens + self._predict_decode(req.num_prefill_tokens)
            return (-req.num_restarts, predicted_total, req.arrived_at)

        self._request_queue.sort(key=_priority)

        # ---------- projected replica states (without unrouted) ----------
        proj_blocks: Dict[int, int] = {}
        proj_backlog_tokens: Dict[int, int] = {}
        proj_queue_len: Dict[int, int] = {}
        blk_size: Dict[int, int] = {}
        token_capacity: Dict[int, int] = {}

        for rid, rep in replicas.items():
            bs = rep.block_size
            blk_size[rid] = bs
            token_capacity[rid] = rep.num_blocks * bs

            blocks = rep.num_allocated_blocks  # currently allocated blocks
            backlog_tokens = 0
            qlen = len(rep.active_queue) + len(rep.pending_queue)

            # active requests
            for rq in rep.active_queue:
                # remaining future blocks for this request
                total_tokens_goal = rq.num_prefill_tokens + self._predict_decode(rq.num_prefill_tokens)
                future_blocks = self._ceil_div(total_tokens_goal, bs)
                already_blocks = self._ceil_div(rq.num_processed_tokens, bs)
                blocks += max(0, future_blocks - already_blocks)

                # backlog tokens (remaining prefill)
                if rq.num_processed_tokens < rq.num_prefill_tokens:
                    backlog_tokens += rq.num_prefill_tokens - rq.num_processed_tokens

            # pending requests
            for rq in rep.pending_queue:
                total_tokens_goal = rq.num_prefill_tokens + self._predict_decode(rq.num_prefill_tokens)
                blocks += self._ceil_div(total_tokens_goal, bs)
                backlog_tokens += rq.num_prefill_tokens

            proj_blocks[rid] = blocks
            proj_backlog_tokens[rid] = backlog_tokens
            proj_queue_len[rid] = qlen

        avg_queue_len = (sum(proj_queue_len.values()) / num_repls) + 1e-6

        # ---------- greedy assignment loop ----------
        mapping: List[Tuple[int, 'Request']] = []  # type: ignore[name-defined]
        remaining: List['Request'] = []            # requests we skip this tick

        while self._request_queue:
            req = self._request_queue.pop(0)
            pred_decode = self._predict_decode(req.num_prefill_tokens)
            total_tokens_req = req.num_prefill_tokens + pred_decode

            best_rid: int | None = None
            best_cost: float = float('inf')
            best_util_after: float = 0.0

            for rid, rep in replicas.items():
                bs = blk_size[rid]
                req_blocks = self._ceil_div(total_tokens_req, bs)
                pf_blocks = self._ceil_div(req.num_prefill_tokens, bs)

                free_now_blocks = rep.num_blocks - rep.num_allocated_blocks
                deficit_blocks = max(0, pf_blocks - free_now_blocks)

                util_after = (proj_blocks[rid] + req_blocks) / rep.num_blocks
                backlog_after = proj_backlog_tokens[rid] + req.num_prefill_tokens
                queue_after = proj_queue_len[rid] + 1

                cost = (
                    self._W_UTIL * (util_after ** 2) +
                    self._W_BACKLOG * (backlog_after / (token_capacity[rid] + 1e-6)) +
                    self._W_QUEUE * (queue_after / avg_queue_len) +
                    self._W_DEFICIT * (deficit_blocks / (rep.num_blocks + 1e-6))
                )

                if util_after > 1.0:
                    cost += self._OVER_CAP_PEN * (util_after - 1.0) ** 2

                # discount for restarts
                if req.num_restarts:
                    cost *= (1.0 - self._RESTART_DISCOUNT) ** req.num_restarts

                if cost < best_cost - 1e-12:
                    best_cost = cost
                    best_rid = rid
                    best_util_after = util_after

            if best_rid is None:
                remaining.append(req)
                continue

            # ------ admission control ------
            cap = self._SOFT_UTIL_CAP_RESTART if req.num_restarts else self._SOFT_UTIL_CAP
            if best_util_after > cap:
                # keep for next tick
                remaining.append(req)
                continue

            # commit placement
            sel = best_rid
            mapping.append((sel, req))

            bs_sel = blk_size[sel]
            req_blocks_sel = self._ceil_div(total_tokens_req, bs_sel)
            proj_blocks[sel] += req_blocks_sel
            proj_backlog_tokens[sel] += req.num_prefill_tokens
            proj_queue_len[sel] += 1
            avg_queue_len = (sum(proj_queue_len.values()) / num_repls) + 1e-6

        # push remaining requests back to queue (maintain order)
        self._request_queue = remaining + self._request_queue

        return mapping
\end{lstlisting}

\tikzset{
  pattern prompt/.style={
    rectangle split,
    rectangle split parts=2,
    rectangle split part fill={teal!30,teal!5}, 
    draw,
    thick,
    rounded corners,
    text width=0.45\textwidth,
    font=\scriptsize,
    align=left,
    inner sep=4pt,
  },
  pattern prompt title/.style={
    rectangle split part fill=gray!60,
    text=white,
    font=\bfseries\footnotesize
  }
}
\newpage
\begin{figure*}[t]
\centering
\begin{tikzpicture}
\node[pattern prompt, text width=\textwidth] (evolveprompt) {
\textbf{Prompt for Using LLM as is}
\nodepart{two}
\begin{varwidth}{\textwidth}
  \textbf{System Instruction:} You are an AI expert in designing system algorithms and optimization techniques. Your task is to create efficient algorithms for various system optimization problems.\\[2pt]
  \textbf{Objective:} Please implement a scheduler for a LLM inference. Design a solution that \emph{minimizes the average request response time} across all requests.\\[2pt]
  \textbf{System Model:} Here is how the load balance works:
\begin{verbatim}
1. The load balancer manages a number (e.g., 16) of LLM serving node called `replica_scheduler`s.
The load balancer routes requests to any of these replicas, and must eventually route all requests.

2. The load balancer makes routing decisions per each request. The load balancer knows these three key properties per request:
`_arrived_at`: When the request was received at the load balancer. `_num_prefill_tokens`: number of tokens to prefill.
`num_processed_tokens`: number of tokens that have been processed so far. A request has some number of decode tokens but this is
not known until the request is completed.

3. Each replica maintains two queues: `pending_queue` and `active_queue`. `pending_queue` contains requests where the prefill has not
started. `num_processed_tokens` is 0 and there is no memory allocated in the GPU for these requests. `active_queue` contains requests
that are currently being processed. `num_processed_tokens > 0` and some memory is allocated in the GPU for these requests.
If `num_processed_tokens < num_prefill_tokens`, the request is still in the prefill phase. Otherwise, the request is in the decode phase.
A request cannot be in both queues at the same time.

4. Each replica has to allocate memories for requests currently being processed, in the `active_queue`, in blocks (16 tokens at a time).
In case there is no memery left at a replica for continuing decoding of active requests, replicas will evict newer requests to free memory
for earlier requests.
This removes the request from the 'active_queue', frees its allocated memory, resets its state, and adds it to the 'pending_queue'.
If the evicted request was in the decode phase, the 'num_prefill_tokens' is updated to 'num_processed_tokens'.

5. The load balancer can observe the state of all replicas, meaning all requests in `pending_queue` and `active_queue` of each replica.\end{verbatim}

\textbf{Implementation:} Please implement the following according to the specifications:\\
\vspace{-10 pt}
\begin{verbatim}
Your task is to implement a custom load balancer by inheriting from BaseGlobalScheduler. To achieve this, you will implement the `schedule`
function of this class. This function is called every time 1) a new request has arrived, or 2) a replica has finished a request.
To return the routing decisions, this function should return a list of tuples.
Each tuple consists of 1) the id number of the replica to route to and 2) the request to be routed `(replica_id, request)`.
Note that routed requests should be popped from `self._request_queue`. Do not change the properties of requests or replicas.

Here is how you can access some helpful metrics to guide the decision-making process.

1. The CustomGlobalScheduler you will implement can see the following elements from a parent class (BaseGlobalScheduler):
<Some elements>
2. Each replica in `_replica_schedulers` is a `ReplicaScheduler` object, and has the following READ-ONLY properties:
<Some properties>
3. Each request has the following READ-ONLY properties:
<Some properties>

Now, implement the load balancer, i.e., CustomGlobalScheduler. Only output the full code for the CustomGlobalScheduler class.

\end{verbatim}

  \textbf{Guidelines:}\\
  To design the algorithm, first consider the requirements and system model carefully to develop an overall strategy. Then, implement your solution.

  \end{varwidth}
};

\end{tikzpicture}
\caption{Prompt for using an LLM as-is for the request-routing problem.}
\label{fig:llm_as_is_prompt}
\end{figure*}

\begin{figure*}[t]
\centering
\begin{tikzpicture}
\node[pattern prompt, text width=\textwidth] (evolveprompt) {
\textbf{Basic Prompt for FunSearch}
\nodepart{two}
\begin{varwidth}{\textwidth}

  \textbf{System Instruction:} You are an AI expert in designing system algorithms and optimization techniques. Your task is to create efficient algorithms for various system optimization problems.\\[2pt]
  \textbf{Objective:} Please implement a scheduler for a LLM inference. Design a solution that \emph{minimizes the average request response time} across all requests.\\[2pt]
  \textbf{System Model:} Here is how the load balance works: \vspace{-5 pt}
\begin{verbatim}
1. The load balancer manages a number (e.g., 16) of LLM serving node called `replica_scheduler`s.
The load balancer routes requests to any of these replicas, and must eventually route all requests.

2. The load balancer makes routing decisions per each request. The load balancer knows these three key properties per request:
`_arrived_at`: When the request was received at the load balancer. `_num_prefill_tokens`: number of tokens to prefill.
`num_processed_tokens`: number of tokens that have been processed so far. A request has some number of decode tokens but this is
not known until the request is completed.

3. Each replica maintains two queues: `pending_queue` and `active_queue`. `pending_queue` contains requests where the prefill has not
started. `num_processed_tokens` is 0 and there is no memory allocated in the GPU for these requests. `active_queue` contains requests
that are currently being processed. `num_processed_tokens > 0` and some memory is allocated in the GPU for these requests.
If `num_processed_tokens < num_prefill_tokens`, the request is still in the prefill phase. Otherwise, the request is in the decode phase.
A request cannot be in both queues at the same time.

4. Each replica has to allocate memories for requests currently being processed, in the `active_queue`, in blocks (16 tokens at a time).
In case there is no memery left at a replica for continuing decoding of active requests, replicas will evict newer requests to free memory
for earlier requests.
This removes the request from the 'active_queue', frees its allocated memory, resets its state, and adds it to the 'pending_queue'.
If the evicted request was in the decode phase, the 'num_prefill_tokens' is updated to 'num_processed_tokens'.

5. The load balancer can observe the state of all replicas, meaning all requests in `pending_queue` and `active_queue` of each replica.\end{verbatim}

\textbf{Implementation:} Please implement the following according to the specifications:\\
\vspace{-10 pt}
\begin{verbatim}
Your task is to implement a custom load balancer by inheriting from BaseGlobalScheduler. To achieve this, you will implement the `schedule`
function of this class. This function is called every time 1) a new request has arrived, or 2) a replica has finished a request.
To return the routing decisions, this function should return a list of tuples.
Each tuple consists of 1) the id number of the replica to route to and 2) the request to be routed `(replica_id, request)`.
Note that routed requests should be popped from `self._request_queue`. Do not change the properties of requests or replicas.

Here is how you can access some helpful metrics to guide the decision-making process.

1. The CustomGlobalScheduler you will implement can see the following elements from a parent class (BaseGlobalScheduler):
<Some elements>
2. Each replica in `_replica_schedulers` is a `ReplicaScheduler` object, and has the following READ-ONLY properties:
<Some properties>
3. Each request has the following READ-ONLY properties:
<Some properties>

Now, implement the load balancer, i.e., CustomGlobalScheduler. Only output the full code for the CustomGlobalScheduler class.

\end{verbatim}

  \textbf{Guidelines:}\\
  To design the algorithm, first consider the requirements and system model carefully to develop an overall strategy. Then, implement your solution.

\begin{verbatim}
Some other implementations and their achieved average request response time:
# Implementation #1 (response time: 54.054 s):
<algorithm 1> 
# Implementation #2 (response time: 55.55 s):
<algorithm 2>
# Implementation #3 (response time: 57.142 s):
<algorithm 3>
\end{verbatim}
  \end{varwidth}
};

\end{tikzpicture}
\caption{Base prompt for our FunSearch evaluation.}
\label{fig:funsearch_prompt}
\end{figure*}


\begin{figure*}[t]
\centering
\begin{tikzpicture}
\node[pattern prompt, text width=\textwidth] (evolveprompt) {
\textbf{System Prompt for Openevolve}
\nodepart{two}
\begin{varwidth}{\textwidth}
\begin{Verbatim}[breaklines=true, breakanywhere=true]
Please implement a scheduler for a LLM inference cluster.

System Model:
Here is how the load balance works:

1. The load balancer manages a number of LLM serving nodes called `replica_scheduler`s.
  - The load balancer routes requests to any of these replicas.
  - The load balancer must eventually route all requests.

2. The load balancer makes routing decisions per each request. The load balancer knows these three key properties per request:
  - `_arrived_at`: When the request was received at the load balancer.
  - `_num_prefill_tokens`: number of tokens to prefill.
  - `num_processed_tokens`: number of tokens that have been processed so far.
  - A request has some number of decode tokens but this is not known until the request is completed.

3. Each replica maintains two queues: `pending_queue` and `active_queue`.
  - `pending_queue` contains requests where the prefill has not started. `num_processed_tokens` is 0 and there is no memory allocated in the GPU for these requests.
  - `active_queue` contains requests that are currently being processed. `num_processed_tokens > 0` and some memory is allocated in the GPU for these requests. If `num_processed_tokens < num_prefill_tokens`, the request is still in the prefill phase. Otherwise, the request is in the decode phase.
  - A request cannot be in both queues at the same time.

4. Each replica has to allocate memories for requests currently being processed, in the `active_queue`, in blocks (16 tokens at a time).
  - In case there is no memory left at a replica for continuing decoding of active requests, replicas will evict newer requests to free memory for earlier requests.
  - This removes the request from the 'active_queue', frees its allocated memory, resets its state, and adds it to the 'pending_queue'.
  - If the evicted request was in the decode phase, the 'num_prefill_tokens' is updated to 'num_processed_tokens'.

5. The load balancer can observe the current state of all replicas, meaning all requests in `pending_queue` and `active_queue` of each replica.

Objective:
Design a solution that Minimize the average request completion time across all requests.


Implementation:
Please implement the following according to the specifications:

Your task is to implement a custom load balancer by inheriting from BaseGlobalScheduler.
To achieve this, you will implement the `schedule` function of this class.
This function is called everytime 1) a new request has arrived, or 2) a replica has finished a request.
To return the routing decisions, this function should return a list of tuples. Each tuple consists of 1) the id number of the replica to route to and 2) the request to be routed `(replica_id, request)`.
Note that routed requests should be popped from `self._request_queue`.
Do not change the properties of requests or replicas.

Here is how you can access some helpful metrics to guide the decision-making process.

1. The CustomGlobalScheduler you will implement can see the following elements from a parent class (BaseGlobalScheduler):
<Some elements>
2. Each replica in `_replica_schedulers` is a `ReplicaScheduler` object, and has the following READ-ONLY properties:
<Some properties>
3. Each request has the following READ-ONLY properties:
<Some properties>

Now, implement the load balancer, i.e., CustomGlobalScheduler. Only output the full code for the CustomGlobalScheduler class.

<Class signiture>

Guidelines:

To design the algorithm, first consider the requirements and system model carefully to develop an overall strategy. Then, implement your solution.

\end{Verbatim}
\end{varwidth}
};
\end{tikzpicture}
\caption{System prompt used for our OpenEvolve evaluation.}
\label{fig:openevolve_prompt}
\end{figure*}


\begin{figure*}[t]
\centering
\begin{tikzpicture}
\node[pattern prompt, text width=\textwidth] (evolveprompt) {
\textbf{The user's prompt to \name for the LLM request-routing problem.}
\nodepart{two}
\begin{varwidth}{\textwidth}
\begin{verbatim}
Design an efficient request scheduler for a distributed LLM serving cluster. Use the simulator (the current working directory) to evaluate
your ideas.

System overview:
The system has a number of LLM serving instances (replicas) that can process the requests. Incoming requests are first processed by a
`global scheduler'. The global scheduler maintains a queue of requests. It decides when and which replica to send each request.

The base class for the global scheduler "BaseGlobalScheduler" can be found at: "scheduler/global_scheduler/base_global_scheduler.py"
A simple implementation of the global scheduler is LLQGlobalScheduler which dispatches requests to the replica with least loaded queue,
and can be found here: "scheduler/global_scheduler/llq_global_scheduler.py"

The entry point for the global scheduler is the schedule() function. This is called by the simulator after each request
arrival and request completion event.

Each serving instance schedules its incoming requests via a `replica scheduler'. The replica scheduler creates batches of work to be
processed on GPUs.

The base class for replica scheduler “BaseReplicaScheduler” can be found at: "scheduler/replica_scheduler/base_replica_scheduler.py"
For this design task, we will use the vLLM replica scheduler provided here: "scheduler/replica_scheduler/vllm_replica_scheduler.py"

To summarize, the life of a request in the system is as follows:
incoming request → global scheduler → replica scheduler → Batch processing by GPU (prefill / decode)

Every request is first processed in prefill stage to compute the kv-cache of the input tokens. Once prefill is complete, the request enters
the decode stage where its output tokens are computed incrementally. The total number of decode tokens is not known until a request finishes.
Only the number of prefill tokens is known when a request first arrives.

Objective:
Your task is to optimize the global scheduler. The primary performance metric is the average response time of requests.

Evaluation:
A benchmark is provided to evaluate your designs. It consists of a 1000-second simulation of a workload running on a cluster with 4
identical a10 GPUs. The workload generates <target_qps> queries-per-second (qps). The benchmark workload can be found in
“data/processed_traces/sharegpt_<target_qps>.csv”. To run the benchmark, use the following command: ./run_all.sh

As a baseline, I ran the benchmark for the LLQ algorithm. The simulation outputs artifacts in a directory like
“simulator_results/sharegpt_llq_<target_qps>/<folder_time_stamp>/”. This directory contains the following files:

1. “config.json” that specifies the experiment configs,
2. "gs_log.csv" is the log of global scheduling events. For every global scheduling event, it writes 4 (num sarathi instances) lines with
the following information: ['time', 'replica_id', 'num_pending_requests', 'num_active_requests', 'num_allocated_blocks', 'num_blocks',
'memory_usage_percent'].
3. "reqs_log.csv" is the log of where each request was eventually routed. For every request, it writes one line with the following
information: ['time', 'replica_id', 'request_id', 'num_prefill_tokens', 'num_decode_tokens'].
4. "request_metrics.csv" provides per-request information. 

Constraints:

Modify only the global scheduler. Do not change the behavior of replica scheduler.
The global scheduler may not use the num_decode_token property of request objects, since the number of decode tokens of a request is 
not known in a real system.

Implement your ideas in "scheduler/global_scheduler/ai_global_scheduler.py", which is prepopulated with a random acheduler.
Experiment by running “./run_all.sh”
and looking at the output found in “simulator_results/sharegpt_AI_<target_qps>/[YYYY-MM-DD_HH-MM-SS-microseconds]”. Iterate on your
design to reduce the average request completion time ("request_e2e_time" in "request_metrics.csv"). Do not interrupt me until you have
found a solution that is at least better that LLQ's average request time (around <time> seconds on this benchmark). It should be possible
to perform much better than LLQ (at least a <targe_improvement>% improvement is expected).

\end{verbatim}
  \end{varwidth}
};

\end{tikzpicture}
\caption{The user's prompt to \name for the LLM request-routing problem.}
\label{fig:glia_task_prompt}
\end{figure*}

\end{document}